\documentclass[letterpaper,10 pt, journal]{IEEEtran}
\usepackage[utf8]{inputenc}
\usepackage{authblk,graphicx,amsmath,amssymb,pgfplots,subcaption,array,multirow,float,xcolor,standalone,hyperref,cite,soul,cuted}

\usepackage{tikz,tikzscale}
\usetikzlibrary{math,calc,arrows.meta,bending,decorations.pathmorphing}
\pgfplotsset{compat=newest} 
\pgfplotsset{plot coordinates/math parser=false} 
\newlength\figureheight 
\newlength\figurewidth 

\definecolor{myblue}{rgb}{0.00000,0.44700,0.74100}%
\definecolor{mypurple}{rgb}{0.49400,0.18400,0.55600}%

\title{Dynamic Inchworm Crawling: Performance Analysis and Optimization of a Three-link Robot}
\author{Benny~Gamus,~Amir~D.~Gat,~and~Yizhar~Or
\thanks{We thank Mr. Yinon Graitzer and Mr. Shay Tovy for constructing the robot and Ms. Bat-Chen Hoze for conducting the experiments.
This work was supported by Israel Science Foundation under grant no. 1005/19, Israel Ministry of Science and Technology under grants number 3-14418 and 3-15622, and Technion's Center for Security Science and Technology (CSST) under grant no. 2027795.}
\thanks{Benny Gamus, Amir D. Gat and Yizhar Or are with the Faculty of Mechanical Engineering, Technion -- Israel Institute of Technology, Technion City, Haifa, Israel 3200003.}
}

\begin{document}
\maketitle

\begin{abstract}
    Inchworm crawling allows for both quasistatic and dynamic gaits at a wide range of actuation frequencies. This locomotion mechanism is common in nonskeletal animals and exploited extensively in the bio-inspired field of soft robotics. In this work we develop and simulate the hybrid dynamic crawling of a three-link robot, with passive frictional contacts. We fabricate and experimentally test such robot under periodic inputs of joints' angles, with good agreement to the theoretical predictions. This allows to comprehend and exploit the effects of inertia in order to find optimal performance in inputs' parameters. A simple criterion of robustness to uncertainties in friction is proposed. Tuning the inputs according to this criterion improves the robustness of low-frequency actuation, while increasing the frequency allows for gaits with both high advancement velocity and robustness. Finally, the advantages of uneven mass distribution are studied. Time-scaling technique is introduced to shape inputs that achieve similar effect without reassembling the robot. A machine-learning based optimization is applied to these inputs to further improve the robot's performance in traveling distance.
\end{abstract}


\section{Introduction}
In mobile robots, legged locomotion is advantageous in negotiating unstructured terrains, where wheeled and tracked vehicles have limited maneuverability or accessibility \cite{wieber2016modeling}. Bipedal legged robots are also widely studied due to their resemblance to humans and other mammals. Though this robotic configuration has relatively low dimensionality, it usually performs complicated \textit{dynamic} locomotion, where the robot constantly undergoes unsteady motion of falling, followed by the replacement of the support foot in a cyclic pattern (\textit{gait}).

Inchworm crawling is a type of legged locomotion which is characterized by multiple persistent ground contacts, which alternate stick-slip transitions. This allows for crawling gaits to be possible at a wide range of actuation frequencies. For rapid actuation, when the inertial effects are not negligible, dynamic inchworm crawling occurs (which will be the focus of this study). However, for slow actuation, the same locomotion mechanism also allows to remain in static balance while performing \textit{quasistatic} movement -- i.e. transitioning within a continuum of static equilibria -- even on two contacts. Probably for that reason crawling is common in many nonskeletal animals, such as worms and caterpillars, octopi and crawling cells. Most of these creatures generate motion by manipulating fluids in their bodies \cite{kim2013soft} -- which is a rather slow process, compared to skeletal animals with rapidly contracting muscles.

The actual inchworm, crawling cells and some crawling robots actively manipulate the contact interaction by adhesion  \cite{kotay2000inchworm,wu2017regulating}. Many crawling robots utilize directional friction \cite{koh2009omegabot, felton2013robot,wang2019study,vikas2016design}, which do not allow to reverse the movement direction, while others perform crawling with passive frictional contacts \cite{umedachi2013highly, guo2017design}. The latter, which will be the focus of this study, is a more complicated yet more general approach.

\begin{figure}
    \centering
    \includegraphics[width=0.85\linewidth]{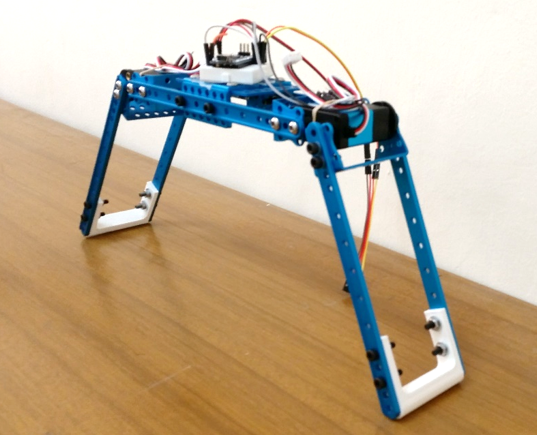}
    \caption{Our three-link robot experimental prototype}
    \label{fig:stiffy}
\end{figure}

In recent years, crawling is exploited extensively \cite{chen2020soft} in the rapidly growing bio-inspired field of \textit{soft robotics} \cite{trivedi2008soft}. In these robots, which draw inspiration from nonskeletal animals, actuation is often similarly created by pressurizing fluids in internal cavities of their compliant structure. Nonetheless, rapid dynamic movement of soft robots can also be achieved -- with other actuation methods, such as magnetic fields \cite{miyashita2015untethered}.

In previous works \cite{gamus2020understanding} we have shown that a three-link model (Fig.\,\ref{fig:3link}) captures well the major phenomena of quasistatic inchworm crawling of a soft bipedal robot. This lumped model is a very basic form of multi-contact bipedal crawling, in analogy to McGeer's biped \cite{mcgeer1990passive} being a benchmark of bipedal walking. Yet, most ``traditional'' articulated legged robots are of higher complexity and often have more legs \cite{yao2015screw,ghanbari2011optimal}. To the best of our knowledge, the minimal three-link robot has not been noticeably studied.

The purpose of this study is to model and analyze the dynamic bipedal frictional inchworm crawling locomotion, in order to comprehend and exploit the effects of inertia, for both soft and articulated robots. First, we develop a model for the hybrid dynamic \cite{goebel2009hybrid} crawling (i.e. with discrete transitions between contact states) of a three-link robot with passive frictional contacts. The robot is actuated in open-loop by prescribing periodic inputs to its joints' angles. We study the effects of frequency and other input parameters on the crawling gait, and find trends and optimal performance. We manufacture and experimentally test a three-link robot (Fig.\,\ref{fig:stiffy}) with good quantitative and excellent qualitative agreement with the theoretical predictions, which proves the applicability of our analysis (see supplementary video \cite{suppVid}). We also investigate the effects of friction uncertainties, which was shown to have major influence \cite{majidi2013influence}, and mass asymmetry, propose a novel input shaping technique and apply machine-learning based optimization to improve the performancein traveling distance. Finally, we discuss a feedback control strategy. 

\section{Problem formulation}
\begin{figure}[b]
    \centering
    \includegraphics[width=\linewidth]{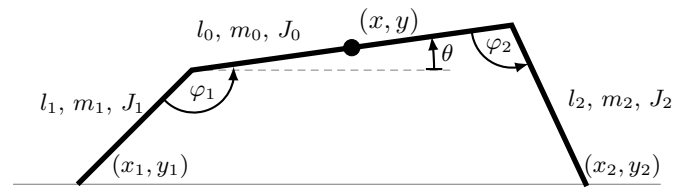}
    \caption{Three-link robot model}
    \label{fig:3link}
\end{figure}

In order to investigate crawling at frequency range where inertial effects are significant, we have manufactured the three-link robot prototype in Fig.\,\ref{fig:stiffy}. The robot has a central link with length $l_0$, mass $m_0$ and moment of inertia $J_0$ and two distal links, with length $l_i$, mass $m_i$ and moment of inertia $J_i$, for $i=1,2$ (see Fig.\,\ref{fig:3link}). For most of this work we consider identical distal links $l_1=l_2\equiv l$, $m_1=m_2\equiv m$, $J_1=J_2\equiv J$, and in Section \ref{sec:learning} we investigate the influence of asymmetric mass distribution. The experimental setup parameters are summarized in Table\,\ref{table}. Two servomotors at the joints receive a sequence of angle commands from the microcontroller (in open-loop) and track it with internal closed-loop control.

Assuming planar motion and point-contacts, a corresponding three-link model is proposed in Fig.\,\ref{fig:3link}. The motion of the robot can be described by the generalized coordinates $\mathbf{q}(t)=[x\quad y\quad \theta\quad \varphi_1\quad \varphi_2]^\text{T}$, where $(x,y,\theta)$ are the planar position and absolute orientation angle of the central link, and $(\varphi_1,\varphi_2)$ are the joint angles. Throughout this work we assume that the two joint angles $\mathbf{q}_\text{c}(t)=[\varphi_1(t)\ \varphi_1(t)]^\text{T}$ are prescribed directly as known periodic input functions (and hence their time-derivatives $\dot{\mathbf{q}}_\text{c}(t),\ \ddot{\mathbf{q}}_\text{c}(t)$ are also known). This assumption, which is quite reasonable for a robot with joints controlled in closed-loop, significantly simplifies the analysis by effectively reducing the number of dynamically-evolving degrees-of-freedom (DoF).

A closed-chain four-bar mechanism has only one DoF. Hence, for general prescribed two joint angles $\boldsymbol{q}_\text{c}(t)$, at least one of the contacts is always constrained to slip (except for discrete times, where the relative velocity between feet vanishes). Therefore, there are only three possible combinations of the contacts' states -- stick-slip, slip-stick and slip-slip -- which we now turn to model.

\section{Hybrid dynamic model}

In this section we introduce the derivation of a hybrid dynamic model which accounts for the different contact states of our three-link robot.


Denoting the position vectors of the contacts as $\mathbf{r}_i~=~[x_i\quad y_i]^\text{T}$ (for $i=1,2$), the velocities of the $i$-th contact are
\begin{equation}\label{eq:v}
    \mathbf{v}_i=\dot{\mathbf{r}}_i=\mathbf{W}_i(\mathbf{q})\dot{\mathbf{q}},
\end{equation}
with the Jacobian matrices
\begin{equation}\label{eq:Jac}
    \mathbf{W}_i(\mathbf{q})=\frac{\partial \mathbf{r}_i(\mathbf{q})}{\partial \mathbf{q}^\text{T}}.
\end{equation}

The dynamic motion equations can be written in a standard matrix form \cite{murray1994mathematical}
\begin{equation}\label{eq:eom}
    \mathbf{M}(\mathbf{q})\ddot{\mathbf{q}}+\mathbf{B}(\mathbf{q},\dot{\mathbf{q}})+\mathbf{G}(\mathbf{q})=\mathbf{E}\boldsymbol{\tau}(t)+\mathbf{W}(\mathbf{q})^\text{T}\mathbf{F}(t),
\end{equation}
where the terms of the matrices $\mathbf{M}, \mathbf{B}, \mathbf{G}, \mathbf{E}$ and the Jacobian $\mathbf{W}=\left[{\mathbf{W}_1}^\text{T}\ {\mathbf{W}_2}^\text{T}\right]^\text{T}$ are given in the Appendix, and $\boldsymbol{\tau}(t)~=~[\tau_1(t)\ \tau_2(t)]^\text{T}$ are the internal input torques at the joints.
The term $\mathbf{F}(t)=[f_1^x\quad f_1^y\quad f_2^x\quad f_2^y]^\text{T}$ is the vector of generalized constraint forces, acting at the $i$-th contact in the normal (superscript $y$) and tangential (superscript $x$) directions (via the Jacobian $\mathbf{W}$). The contact forces vary with the different locomotion modes as we introduce next.

Though the constrained coordinates $\mathbf{q}_\text{c}(t)$ (and their time-derivatives) are known, we are still left with three unconstrained body coordinates $\mathbf{q}_\text{b}(t) = [x\quad y\quad \theta]^\text{T}$, two actuation torques $\boldsymbol{\tau}(t)$ and four contact forces $\mathbf{F}(t)$. This gives 9 unknowns with 5 motion equations (\ref{eq:eom}), hence the motion is governed by additional constraints, as introduced next.

In inchworm crawling, the legs maintain persistent contact with the ground, giving two kinematic constraints $y_1=y_2=0$ (these assumptions are constantly verified by calculating the normal forces at the contacts and requiring $f_i^y>0$). If one of the legs is in contact-sticking (stick-slip or slip-stick modes) we also get a third constraint $\dot{x}_i=0$ for the sticking contact, yet additional constitutive relations are required. Coulomb's dry friction model dictates that the tangential forces $f_i^x$ must maintain
\begin{subequations}\label{eq:ForceStickSlip}
\begin{equation} \label{eq:stick}
    |f_i^x| \leq \mu\, f_i^y \quad \text{-- for a sticking contact}
\end{equation}
and
\begin{equation} \label{eq:slip}
    f_i^x = -\mu\, f_i^y\, \text{sign}\, \dot{x}_i, \quad \text{-- for a slipping contact},
\end{equation}
\end{subequations}
where $\mu$ is Coulomb's friction coefficient (for simplicity, we do not distinguish between static and kinetic friction coefficients). Thus the overall count gives 9 unknowns with 5 equations of motion, two constraints in the normal direction, and two tangential constrains -- either $\dot{x}_i=0$ for a sticking contact or (\ref{eq:slip}) for slippage. This system is complete and allows for a closed solution.

We can now decompose matrices $\mathbf{M}$ and $\mathbf{W}$ into blocks corresponding to the constrained and unconstrained coordinates, and rearrange (\ref{eq:eom}) and the constraints (using (\ref{eq:v})) such that all the unknowns are on one side of the equation (see Appendix for details). These motion equations can be solved simultaneously with the constraints for each contact state with a numerical solver (for our simulation we are using MATLAB\textregistered\ \texttt{ode45} with event detection).

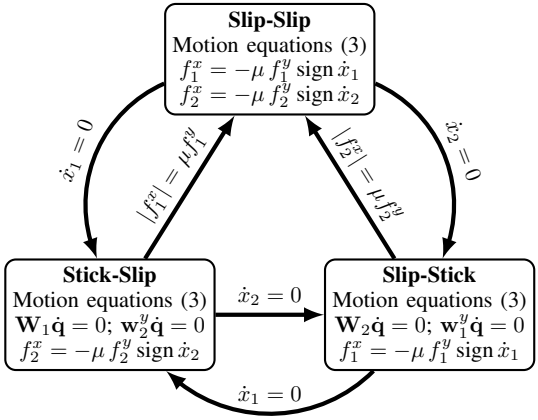
\begin{figure}
    \centering
    \begin{tikzpicture}[scale=0.85, every node/.style={scale=0.85}]

\path (0,4) node [rectangle, rounded corners, thick, draw, align=center] (slip-slip)
{\textbf{Slip-Slip}\\
Motion equations (\ref{eq:eom})\\
$f_1^x = -\mu\, f_1^y\, \text{sign}\, \dot{x}_1$\\
$f_2^x = -\mu\, f_2^y\, \text{sign}\, \dot{x}_2$}
      (-2.5,0) node [rectangle, rounded corners, thick, draw, align=center] (stick-slip)
{\textbf{Stick-Slip}\\
Motion equations (\ref{eq:eom})\\
$\mathbf{W}_1\dot{\mathbf{q}}=0$; $\mathbf{w}_2^y\dot{\mathbf{q}}=0$\\
$f_2^x = -\mu\, f_2^y\, \text{sign}\, \dot{x}_2$}
      (2.5,0) node [rectangle, rounded corners, thick, draw, align=center] (slip-stick)
{\textbf{Slip-Stick}\\
Motion equations (\ref{eq:eom})\\
$\mathbf{W}_2\dot{\mathbf{q}}=0$; $\mathbf{w}_1^y\dot{\mathbf{q}}=0$\\
$f_1^x = -\mu\, f_1^y\, \text{sign}\, \dot{x}_1$};

\draw [-latex,ultra thick] (slip-slip) to [bend left=45] node [right,above,sloped] {$\dot{x}_2=0$} (slip-stick);
\draw [-latex,ultra thick] (slip-stick) to node [above,sloped] {$|f_2^x|=\mu f_2^y$} (slip-slip);

\draw [-latex,ultra thick] (slip-slip) to [bend right=45] node [left,above,sloped] {$\dot{x}_1=0$} (stick-slip);
\draw [-latex,ultra thick] (stick-slip) to node [above,sloped] {$|f_1^x|=\mu f_1^y$} (slip-slip);

\draw [-latex,ultra thick] (slip-stick) to [bend left=45] node [above] {$\dot{x}_1=0$} (stick-slip);
\draw [-latex,ultra thick] (stick-slip) to node [above] {$\dot{x}_2=0$} (slip-stick);

\end{tikzpicture}
    \caption{Transition graph of contacts states}
    \label{fig:hybridStates}
\end{figure}

The conditions for transitions between contact states are as follows. A contact remains in slippage as long as its tangential velocity $\dot{x}_i\neq0$. When $\dot{x}_i=0$, the contact either turns to contact-sticking, or reverses slippage direction. Contact-sticking may only hold as long as the tangential friction force satisfies (\ref{eq:stick}), otherwise slippage occurs. The simulation detects such crossings and switches accordingly the constraints we solve for. Fig.\,\ref{fig:hybridStates} depicts a transition graph of all the contact states. 

\section{Analysis and results}\label{sec:analysis}

\begin{figure*}
\centering
    \begin{subfigure}{0.49\linewidth}
	    \input{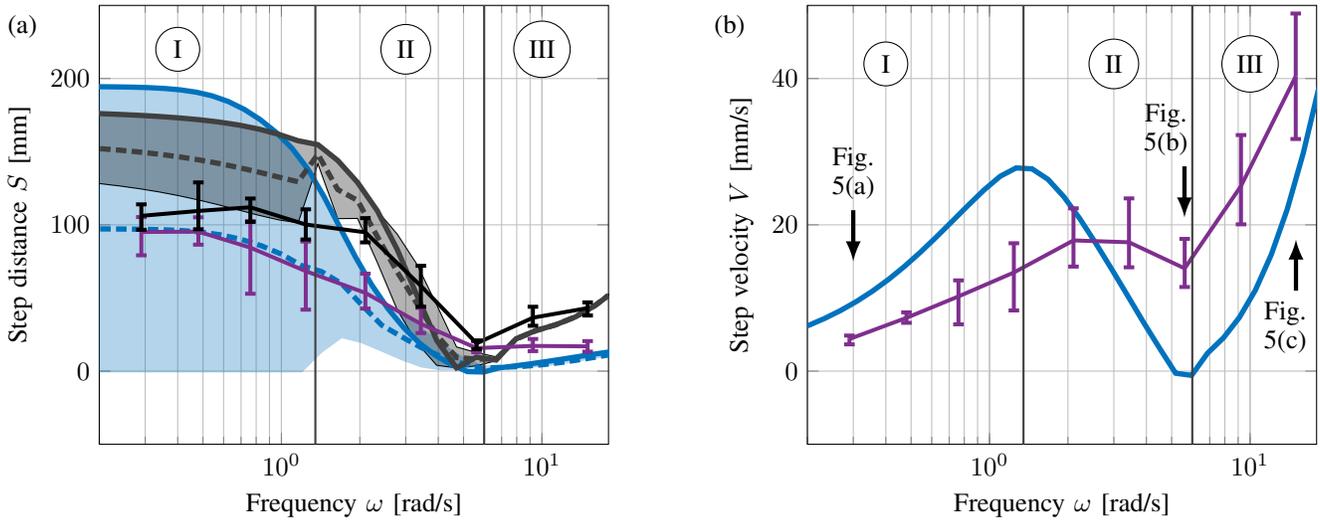}
        \label{fig:S_w}
    \end{subfigure}
    ~
    \begin{subfigure}{0.49\linewidth}
%
%
\definecolor{mycolor1}{rgb}{0.00000,0.44700,0.74100}%
\definecolor{mycolor2}{rgb}{0.49400,0.18400,0.55600}%
\begin{tikzpicture}

\begin{axis}[%
extra description/.code={\node[anchor=north east] at ([xshift=-20pt]current axis.north west) {(b)};},
width=0.75\textwidth,
at={(0.758in,0.567in)},
scale only axis,
xmode=log,
xmin=0.2,
xmax=18,
xminorticks=true,
xlabel={Frequency $\omega$ [rad/s]},
xmajorgrids,
xminorgrids,
ymin=-10,
ymax=50,
ylabel={Step velocity $V$ [mm/s]},
ymajorgrids,
axis background/.style={fill=white},
legend style={at={(0.164,0.81)},anchor=south west,legend cell align=left,align=left,draw=white!15!black}
]
\addplot [color=mycolor1,solid,line width=2.0pt]
  table[row sep=crcr]{%
0.1	3.09646023088773\\
0.115139539932645	3.56525006404097\\
0.132571136559011	4.10501252118524\\
0.152641796717523	4.72538252721446\\
0.175751062485479	5.43981552831523\\
0.202358964772516	6.26142083294019\\
0.232995181051537	7.20541701591776\\
0.268269579527973	8.28831578835139\\
0.308884359647748	9.52714022808324\\
0.355648030622313	10.9376788268727\\
0.409491506238043	12.531147175063\\
0.471486636345739	14.3088249703557\\
0.542867543932386	16.2556409085016\\
0.625055192527397	18.3364765194016\\
0.719685673001152	20.499155493437\\
0.828642772854684	22.6792139141687\\
0.954095476349994	24.7862639018994\\
1.09854114198756	26.6330494554272\\
1.2648552168553	27.7783805786173\\
1.45634847750124	27.6545877923323\\
1.67683293681101	26.2201208964789\\
1.93069772888325	23.8256494378729\\
2.22299648252619	20.8187071419818\\
2.55954792269954	17.4563258892549\\
2.94705170255181	13.9091386637667\\
3.39322177189533	10.2774725991376\\
3.90693993705462	6.61981594508219\\
4.49843266896945	3.10539529362017\\
5.17947467923121	-0.290411190170948\\
5.96362331659464	-0.582617027292496\\
6.866488450043	2.44766944463767\\
7.9060432109077	4.54996854996386\\
9.10298177991522	7.34537608698595\\
10.4811313415469	11.0707532811606\\
12.0679264063933	15.9499289265956\\
13.8949549437314	22.1570314597696\\
15.9985871960606	29.8025091640376\\
18.4206996932672	38.9477816035764\\
};

\addplot [color=mycolor2,solid,line width=1.5pt]
 plot [error bars/.cd, y dir = both, y explicit,error bar style={line width=1.5pt},error mark options={rotate=90,line width=1.5pt,mark size=2pt}]
 table[row sep=crcr, y error plus index=2, y error minus index=3]{%
0.29	4.38979572486635	0.471703420335759	0.736171189271562\\
0.48	7.28420343542987	0.749428796031117	0.680673860615418\\
0.76	10.212463402389	2.19417370744211	3.81621722545747\\
1.24	13.4830974829275	4.00032766362617	5.19036100411289\\
2.1	17.8710310949597	4.38837924587283	3.57955382507982\\
3.43	17.6162399465638	5.99945698830916	3.42280212162861\\
5.61	14.0446756996272	4.03572372296981	2.54464880762477\\
9.23	25.3108721492391	6.96306059125904	5.24433044531535\\
14.97	40.3127375076096	8.57717819310843	8.57717819310843\\
};

\draw [thick,darkgray] (axis cs: 1.35,-10) -- (axis cs: 1.35,50);
\draw (axis cs: 0.4,42) node [circle,draw,fill=white] {I};
\draw [thick,darkgray] (axis cs: 6,-10) -- (axis cs: 6,50);
\draw (axis cs: 3,42) node [circle,draw,fill=white] {II};
\draw (axis cs: 10,42) node [circle,draw,fill=white] {III};

\draw [-latex,ultra thick,black] (axis cs: 0.3,22) -- (axis cs: 0.3,15) node [pos=0,above,align=center] {Fig.\\ 5(a)};
\draw [-latex,ultra thick,black] (axis cs: 5.6,28) -- (axis cs: 5.6,21) node [pos=0,left=7pt,above,align=center] {Fig.\\ 5(b)};
\draw [-latex,ultra thick,black] (axis cs: 15,11) -- (axis cs: 15,18) node [pos=0,left=4pt,below,align=center] {Fig. \\ 5(c)};

\end{axis}
\end{tikzpicture}%
	    \label{fig:V_w}
	\end{subfigure}
\caption{Gait performance versus frequency -- (a) Distance per step. (b) Average velocity. For constant $\psi=20^{\circ}$: \textcolor{myblue}{nominal-friction simulation results (solid blue)}, \colorbox{myblue!30}{$\mu_1/\mu_2 \in [0.9 , 1.1]$ (blue area)}, \textcolor{myblue}{uncertain-friction distance $S_\mu$ (dashed blue)} and \textcolor{mypurple}{experiment results with error-bars (purple)}. For $\psi=\psi^*_\mu(\omega)$ phase of optimal $S_\mu$: \textcolor{gray}{nominal-friction simulation results (gray)}, \colorbox{black!30}{$\mu_1/\mu_2 \in [0.9 , 1.1]$ (gray area)}, \textcolor{gray}{uncertain-friction distance $S_\mu$ (dashed gray)} and experiment results with error-bars (black)}
\label{fig:freq}
\end{figure*}

In this section we investigate the influence of various parameters on the robot's performance via the numerical simulation. Then, in Section \ref{sec:Exp}, these results are compared to experiments (also see supplementary video \cite{suppVid}). We choose concrete periodic functions of the angles as follows
\begin{subequations}\label{eq:ref}
\begin{equation}
    \varphi_1(t)=\varphi_0+A\sin\left(\omega t+\psi/2 \right),
\end{equation}
\begin{equation}
    \varphi_2(t)=\varphi_0+A\sin\left(\omega t-\psi/2 \right),
\end{equation}
\end{subequations}
where $\varphi_0$ is the nominal angle, $A$ is the oscillation amplitude, $\psi$ is the phase difference between the legs, and $\omega$ is the frequency. The performance is measured via the net distance traveled per step (in steady state) $S~\equiv~x_1(t)~-~x_1(t-T)$ and the average step velocity $V\equiv S/T$, where $T\equiv2\pi/\omega$ is the period time.

\begin{figure*}
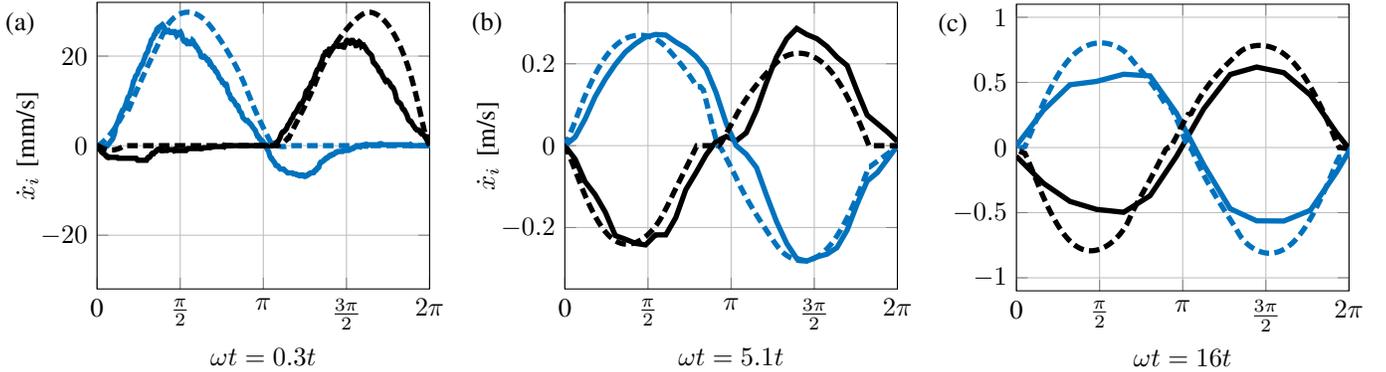

\centering
    \begin{subfigure}{0.32\linewidth}
	    \input{dx_w03.tikz}
        \label{fig:dx_w03}
    \end{subfigure}
    \,
    \begin{subfigure}{0.32\linewidth}
	    \input{dx_w51.tikz}
	    \label{fig:dx_w51}
	\end{subfigure}
	\,
	\begin{subfigure}{0.32\linewidth}
	    \input{dx_w16.tikz}
	    \label{fig:dx_w16}
	\end{subfigure}
\caption{Contacts' velocities \textcolor{myblue}{$\dot{x}_1$ (blue)} and $\dot{x}_2$ (black) -- experiment results (solid curves) and simulation results (dashed curves). (a) $\omega=0.3$ [rad/s] low frequency range {\large\textcircled{\small{I}}}. (b) $\omega=5.1$ [rad/s] frequency range {\large\textcircled{\small{II}}}. (c) $\omega=16$ [rad/s] high frequency range {\large\textcircled{\small{III}}}}
\label{fig:dx}
\end{figure*}

Parametric investigation of the simulation shows some influence and coupling of all the parameters. The most significant effects are achieved by varying the actuation frequency $\omega$, as shown in Fig.\,\ref{fig:freq} (solid blue curves) for $A=18^{\circ}$, $\varphi_0=110^{\circ}$ and $\psi=20^{\circ}$. To further comprehend these results we divide the frequencies into three ranges (marked by {\large\textcircled{\small{I}}} to {\large\textcircled{\small{III}}} in Fig.\,\ref{fig:freq}) and plot in Fig.\,\ref{fig:dx} the contacts' velocities $\dot{x}_i(t)$ for one representative frequency in each range. These selected frequencies are also indicated by arrows in Fig.\,\ref{fig:freq}(b). In the low frequencies (range {\large\textcircled{\small{I}}}), the inertial effects are minor and the robot performs gaits with almost ideal switching -- where the contacts slip consecutively only in the desired advancement direction. This is illustrated by the contacts' velocities $\dot{x}_i(t)$ in Fig.\,\ref{fig:dx}(a) (dashed curves), and corresponds to results achieved with quasistatic analysis in \cite{gamus2020understanding}. In this frequency range, since the distance $S$ is almost steady but the cycle time shortens as $1/\omega$, the average velocity $V$ increases with frequency (see Fig.\,\ref{fig:freq}(b)). In range {\large\textcircled{\small{II}}}, as the frequency increases, a growing portion of the cycle exhibits slippage in the opposing direction, until the advancement is almost entirely canceled. The velocities $\dot{x}_i(t)$ at a frequency with minimal advancement are depicted in Fig.\,\ref{fig:dx}(b). In this range, $S$ shortens faster than the cycle time $T$, and the average velocity decreases. Finally, for even higher frequencies (in range {\large\textcircled{\small{III}}}), though the robot continues to exhibit opposing slippage throughout the cycle, it also develops a rigid-body progression in the desired direction. This increases the net distance only slightly, but since the frequency is in a high range the average velocity rises rapidly. In Fig.\,\ref{fig:dx}(c) we notice that for a short portion of the cycle both contacts have positive slippage velocities $\dot{x}_i>0$. 

\begin{figure}
    \centering
    \begin{subfigure}{0.9\linewidth}
	    \input{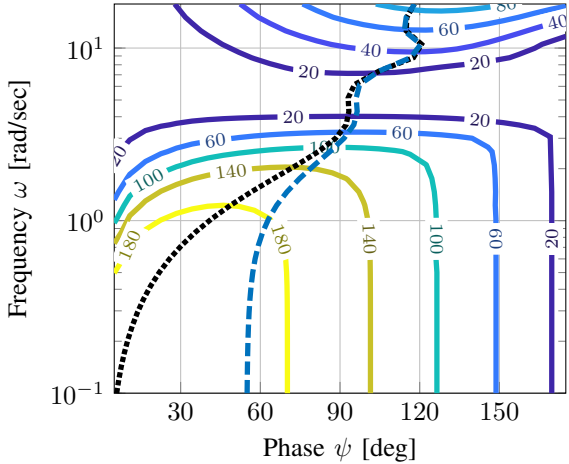}
    \end{subfigure}
    \caption{Step distance $S$ [mm] (contour plot) in parametric space $\psi-\omega$. Phase $\psi^*(\omega)$ of optimal nominal-friction distance $S$ (dotted black curve), \textcolor{myblue}{phase $\psi^*_\mu(\omega)$ of optimal uncertain-friction distance $S_\mu$ (dashed blue curve)}}
    \label{fig:psi_w_map}
\end{figure}

We can also find \textit{optimal phase difference} $\psi^*$ which maximizes the distance $S$ for each frequency. The optimum $\psi^*(\omega)$ shifts (almost monotonically in 
$\omega$) from $\psi^*\rightarrow 0$ for low frequencies (with agreement to the quasistatic limit in \cite{gamus2020understanding}) towards $\psi^*\approx 120^{\circ}$ as the frequency rises. This is depicted by the dotted black curve over the $\psi-\omega$ parametric space in Fig.\,\ref{fig:psi_w_map}. Moving average is applied to the curve in order to reduce numerical noise, especially for the frequency range where $S$ is small.

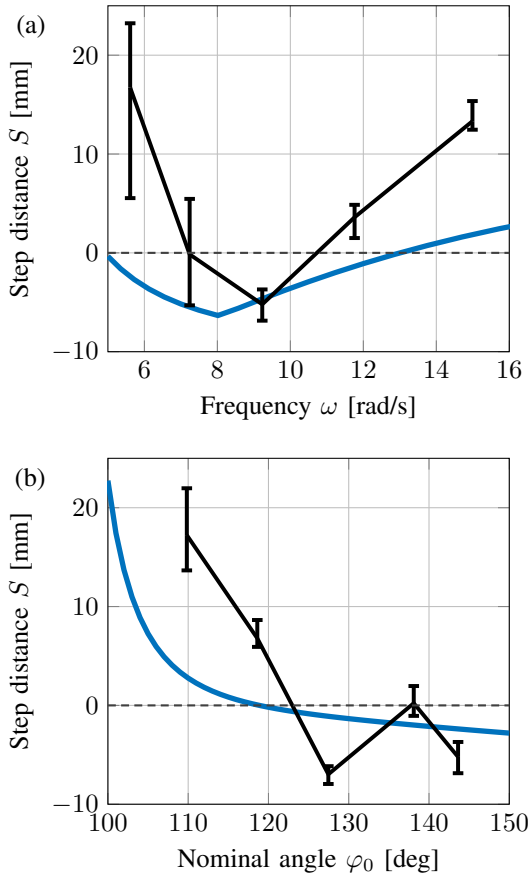
\begin{figure}
\centering
    \begin{subfigure}{0.8\linewidth}
%
%
\definecolor{mycolor1}{rgb}{0.00000,0.44700,0.74100}%
\definecolor{mycolor2}{rgb}{0.49400,0.18400,0.55600}%
\begin{tikzpicture}

\begin{axis}[%
extra description/.code={\node[anchor=north east] at ([xshift=-20pt]current axis.north west) {(a)};},
width=0.75\textwidth,
at={(0.758in,0.515in)},
scale only axis,
unbounded coords=jump,
xmin=5,
xmax=16,
xminorticks=true,
xlabel={Frequency $\omega$ [rad/s]},
xmajorgrids,
xminorgrids,
ymin=-10,
ymax=25,
ylabel={Step distance $S$ [mm]},
ymajorgrids,
axis background/.style={fill=white},
legend style={legend cell align=left,align=left,draw=white!15!black}
]
\addplot [color=mycolor1,solid,line width=2.0pt]
  table[row sep=crcr]{%
5	-0.3457444887974\\
5.34871038997321	-1.6069422973375\\
5.72174056716148	-2.69868054120761\\
6.12078664406156	-3.64485260141413\\
6.54766302358027	-4.46573092978502\\
7.00431064885344	-5.1785692408678\\
7.49280582842448	-5.79809031432636\\
8.01536967690918	-6.33689335914568\\
8.57437821407207	-5.60640395023774\\
9.17237316823345	-4.73599574511766\\
9.81207353312835	-3.83877246574189\\
10.496387930765	-2.92348912635362\\
11.2284278364944	-1.99930999873473\\
12.0115217264244	-1.07549823942859\\
12.849230211503	-0.161097790464731\\
13.7453622270848	0.735368637933458\\
14.7039923515907	1.60616497116576\\
15.729479333008	2.44455243280736\\
16.8264859074658	3.24493414027357\\
18	nan\\
};

\addplot [color=black,solid,line width=1.5pt]
 plot [error bars/.cd, y dir = both, y explicit,error bar style={line width=1.5pt},error mark options={rotate=90,line width=1.5pt,mark size=2pt}]
 table[row sep=crcr, y error plus index=2, y error minus index=3]{%
5.61	16.72	6.51	11.18\\
7.24	-0.15	5.61	5.16\\
9.23	-5.21	1.51	1.65\\
11.76	3.6	1.26	2.1\\
14.99	13.32	2.04	0.86\\
};

\draw [thick,darkgray,densely dashed] (axis cs: 5,0) -- (axis cs: 16,0);

\end{axis}
\end{tikzpicture}%
        \label{fig:S_j0_soft}
    \end{subfigure}
    ~
    \begin{subfigure}{0.8\linewidth}
%
%
\definecolor{mycolor1}{rgb}{0.00000,0.44700,0.74100}%
\definecolor{mycolor2}{rgb}{0.49400,0.18400,0.55600}%
\begin{tikzpicture}

\begin{axis}[%
extra description/.code={\node[anchor=north east] at ([xshift=-20pt]current axis.north west) {(b)};},
width=0.75\textwidth,
at={(0.758in,0.515in)},
scale only axis,
xmin=100,
xmax=150,
xlabel={Nominal angle $\varphi_0$ [deg]},
xmajorgrids,
ymin=-10,
ymax=25,
ylabel={Step distance $S$ [mm]},
ymajorgrids,
axis background/.style={fill=white},
legend style={legend cell align=left,align=left,draw=white!15!black}
]
\addplot [color=mycolor1,solid,line width=2.0pt]
  table[row sep=crcr]{%
100	22.7457385358943\\
101	17.4331768463417\\
102	13.713900369168\\
103	10.9759023167921\\
104	8.88871379945631\\
105	7.25729463543842\\
106	5.96386620434729\\
107	4.92088103815624\\
108	4.06678121263007\\
109	3.35804878631178\\
110	2.76306983097174\\
111	2.25835485526177\\
112	1.82611674495078\\
113	1.45266432965076\\
114	1.12731152422322\\
115	0.841613869722221\\
116	0.588826673290859\\
117	0.363511436260296\\
118	0.161244187831217\\
119	-0.0215989992397535\\
120	-0.188008988282064\\
121	-0.340468240917384\\
122	-0.481048292874904\\
123	-0.611487317090881\\
124	-0.733249591080918\\
125	-0.847573172687765\\
126	-0.955508836909007\\
127	-1.05795134150571\\
128	-1.15566375706963\\
129	-1.24929843650619\\
130	-1.33941377318099\\
131	-1.42648815990318\\
132	-1.51093145642339\\
133	-1.59309441194248\\
134	-1.67327655831834\\
135	-1.75173284385667\\
136	-1.82867914384137\\
137	-1.90429678853782\\
138	-1.97873633778332\\
139	-2.0521206607534\\
140	-2.12454740703663\\
141	-2.19609093334942\\
142	-2.26680370580549\\
143	-2.33671717819662\\
144	-2.40584213000424\\
145	-2.47416839661615\\
146	-2.54166386231164\\
147	-2.60827250630865\\
148	-2.67391118325037\\
149	-2.73846460694123\\
150	-2.80177768752694\\
};

\addplot [color=black,solid,line width=1.5pt]
 plot [error bars/.cd, y dir = both, y explicit,error bar style={line width=1.5pt},error mark options={rotate=90,line width=1.5pt,mark size=2pt}]
 table[row sep=crcr, y error plus index=2, y error minus index=3]{%
143.62	-5.21	1.51	1.65\\
138.1	0.22	1.74	1.28\\
127.46	-6.97	0.82	0.98\\
118.61	6.79	1.85	0.87\\
109.83	17.23	4.74	3.57\\
};

\draw [thick,darkgray,densely dashed] (axis cs: 100,0) -- (axis cs: 150,0);

\end{axis}
\end{tikzpicture}%
	    \label{fig:S_w_145}
	\end{subfigure}
\caption{Direction reversal -- \textcolor{myblue}{simulation results (blue)} and experiment results with error-bars (black). (a) Step distance versus frequency for $\varphi_0=145^{\circ}$. (b) Step distance versus nominal angle for $\omega=9$ [rad/s]}
\label{fig:reverse}
\end{figure}

Another interesting effect of inertia is observed with a variation of the nominal angle. At large enough $\varphi_0$ (i.e. ``flatter'' robot), by varying the input frequency $\omega$ a \textit{direction reversal} occurs, as shown in Fig.\,\ref{fig:reverse}(a) in solid blue curve (for $\varphi_0=145^{\circ}$, $A=18^{\circ}$ and $\psi=20^{\circ}$). The same phenomenon is observed for a sweep of the nominal angle $\varphi_0$ at a frequency in the above range ($\omega=9$ [rad/s]) in Fig.\,\ref{fig:reverse}(b).

\section{Experiments}\label{sec:Exp}

\begin{table}
\normalsize
\centering
\begin{tabular}{ |m{3cm}|c|c|c| } 
    \hline
    Parameter & Notation & Value & Units\\
    \hline
    \hline
    Central link's mass & $m_0$ & 194 & gr \\
    \hline
    Central link's length & $l_0$ & 187 & mm \\
    \hline
    Central link's moment of inertia & $J_0$ & 776.3 & kg mm$^2$ \\
    \hline
    Distal links' mass & $m_1=m_2=m$ & 21 & gr \\
    \hline
    Distal links' length & $l_1=l_2=l$ & 170 & mm \\
    \hline
    Distal links' moment of inertia & $J_1=J_2=J$ & 98.7 & kg mm$^2$ \\
    \hline
    Friction coefficient -- hard tips & $\mu$ & 0.172 & -- \\
    \hline
    Friction coefficient -- soft tips & $\mu$ & 0.398 & -- \\
    \hline
\end{tabular}
\caption{Summary of experimental setup parameters' values}
\label{table}
\end{table}

In order to measure the performance of our robot in Fig.\,\ref{fig:stiffy}, we video it with a simple webcam. The video lens distortion is then corrected with MATLAB\textregistered\ Computer Vision System Toolbox and post-processed with Kinovea software, which gives planar x-y position of selected points on the robot's links. Since the robot is controlled in open-loop, as the frequency increases the servo struggles to follow the prescribed amplitude. Therefore we initially calibrate the amplitudes of inputs' reference trajectories to achieve the desired angles in (\ref{eq:ref}), and validate the other parameters (such as frequency).

In an additional preliminary calibration experiment, we measure the friction coefficient by tilting the plane with the robot rigidly standing on top, and finding (with image processing) the angle at which the robot starts to slip (see Table.\,\ref{table}).

At each set of parameters the robot performs five cycles, and the overall distance traveled is averaged. Each experiment is repeated four times, giving the black curve and error bars (representing the range of observations) in Fig.\,\ref{fig:freq}. The qualitative performance in frequency is very similar to the analytical prediction, giving multiple local extrema in average velocity $V$ in Fig.\,\ref{fig:freq}(b). Yet, the actual distance traveled at the low frequencies range is about two times smaller than the simulation predicts. The differences can be explained by sensitivity analysis to friction variations, which we study in the next section \ref{sec:fric}. We will show that robustness to friction uncertainties has major influence and must be considered in order to improve the predictive power of the model and simulations.

By differentiating the measured horizontal position of the contacts, and filtering with moving average, we get the contacts' velocities $\dot{x}_i(t)$ for selected frequencies in solid curves in Fig.\,\ref{fig:dx}. It is observed that at low frequencies (Fig.\,\ref{fig:dx}(a)) slippage occurs in the undesired direction, resulting in a smaller net traveling distance than predicted. This small but significant divergence from the analytical model can also be explained by friction uncertainties (see Section \ref{sec:fric}). For higher frequencies (Fig.\,\ref{fig:dx}(b) and (c)), the experimental and analytical results match both qualitatively and quantitatively.

Finally, the phenomenon of direction reversal with variation of the nominal angle $\varphi_0$ is also achieved in experiments (black curves in Fig.\,\ref{fig:reverse}) -- see supplementary video \cite{suppVid}. It is of note that this experiment required switching the robot's tips to softer material, with higher friction coefficient, which exhibits more robust behavior at this frequency range. The friction coefficient was measured (see Table\,\ref{table}) and simulated in the corresponding blue curves in Fig.\,\ref{fig:reverse}).


\section{Robustness to friction uncertainties}\label{sec:fric}

\begin{figure}
    \centering
    \begin{subfigure}{0.8\linewidth}
%
%
\definecolor{mycolor1}{rgb}{0.00000,0.44700,0.74100}%
\definecolor{mycolor2}{rgb}{0.49400,0.18400,0.55600}%
\begin{tikzpicture}

\begin{axis}[%
width=0.75\textwidth,
at={(0.758in,0.515in)},
scale only axis,
xmin=5,
xmax=170,
xlabel={Phase $\psi$ [deg]},
xmajorgrids,
ymin=0,
ymax=105,
ylabel={Normalized step distance [\%]},
ymajorgrids,
axis background/.style={fill=white},
legend style={legend cell align=left,align=left,draw=white!15!black}
]

\addplot[area legend,solid,draw=black,fill=black,fill opacity=0.3,forget plot] 
table[row sep=crcr] {%
x	y\\
5	6.45568651835702\\
10.6896551724138	10.3265654712407\\
16.3793103448276	15.1873519936888\\
22.0689655172414	20.2794236597704\\
27.7586206896552	25.6528944060518\\
33.448275862069	31.3214139100747\\
39.1379310344828	37.2074052003763\\
44.8275862068966	43.3787304398405\\
50.5172413793103	49.6507805431484\\
56.2068965517241	56.0003309825131\\
61.8965517241379	62.3898719162942\\
67.5862068965517	68.6175585217403\\
73.2758620689655	74.6285452694065\\
78.9655172413793	80.2893391555062\\
84.6551724137931	85.4698289776398\\
90.3448275862069	90.1034910480368\\
96.0344827586207	93.9566524996283\\
101.724137931034	96.9693976750291\\
107.413793103448	99.0336867308899\\
113.103448275862	100\\
118.793103448276	99.7946853786165\\
124.48275862069	98.3725284660998\\
130.172413793103	95.6135141714207\\
135.862068965517	91.4793445909217\\
141.551724137931	85.838303797511\\
147.241379310345	78.5576349402054\\
152.931034482759	69.5343748246832\\
158.620689655172	58.8521769403945\\
164.310344827586	46.8088571718135\\
170	33.9066170835904\\
170	25.9502809759701\\
164.310344827586	39.3682798075834\\
158.620689655172	51.9571803109997\\
152.931034482759	63.1187606213234\\
147.241379310345	72.6076399556599\\
141.551724137931	80.3550112249908\\
135.862068965517	86.4471585407162\\
130.172413793103	90.9405279165819\\
124.48275862069	93.9779705657038\\
118.793103448276	95.683116106984\\
113.103448275862	96.1752797653803\\
107.413793103448	95.4396739564353\\
101.724137931034	93.5952843767277\\
96.0344827586207	90.7639437291703\\
90.3448275862069	87.040896018436\\
84.6551724137931	82.6005317231647\\
78.9655172413793	77.5334781536955\\
73.2758620689655	71.976648301901\\
67.5862068965517	65.9916219405146\\
61.8965517241379	59.8206651515077\\
56.2068965517241	53.5176063056429\\
50.5172413793103	47.1515811443504\\
44.8275862068966	40.8712260926217\\
39.1379310344828	34.7019949644158\\
33.448275862069	28.7780048480006\\
27.7586206896552	23.0699005752701\\
22.0689655172414	17.6356593032006\\
16.3793103448276	12.4821231849445\\
10.6896551724138	7.5714776579411\\
5	2.81120292084433\\
}--cycle;

\addplot[area legend,solid,draw=mycolor1,line width=1pt,fill=mycolor1,forget plot,opacity=0.3] 
table[row sep=crcr] {%
x	y\\
5	96.732682514541\\
10.6896551724138	99.8931016499795\\
16.3793103448276	100.000002704831\\
22.0689655172414	99.4212157679355\\
27.7586206896552	98.4756386717985\\
33.448275862069	97.2431171723823\\
39.1379310344828	95.7525466959209\\
44.8275862068966	94.0177181318897\\
50.5172413793103	92.0470771332362\\
56.2068965517241	89.8470149897995\\
61.8965517241379	87.4234224328456\\
67.5862068965517	84.7819232159755\\
73.2758620689655	81.9283816466426\\
78.9655172413793	78.6530764573214\\
84.6551724137931	75.6106589079739\\
90.3448275862069	72.160305362358\\
96.0344827586207	68.4220389930304\\
101.724137931034	64.7151718788555\\
107.413793103448	60.7374547894057\\
113.103448275862	56.3912342151743\\
118.793103448276	52.2606774716143\\
124.48275862069	47.6616903509351\\
130.172413793103	43.1929851047552\\
135.862068965517	38.6717219348492\\
141.551724137931	33.7611402932013\\
147.241379310345	29.0162715585455\\
152.931034482759	24.1109877708257\\
158.620689655172	19.0900895870131\\
164.310344827586	14.0105341071253\\
170	8.95773780072988\\
170	8.545035580973\\
164.310344827586	13.4150550597158\\
158.620689655172	18.0853021524887\\
152.931034482759	22.8648422522377\\
147.241379310345	27.4689126960627\\
141.551724137931	31.8636500033121\\
135.862068965517	36.0985807214437\\
130.172413793103	40.4544080703661\\
124.48275862069	44.2983947267841\\
118.793103448276	47.8926300832178\\
113.103448275862	51.6047628940018\\
107.413793103448	54.8802093939678\\
101.724137931034	57.6556462426331\\
96.0344827586207	60.12174496874\\
90.3448275862069	62.4448586051651\\
84.6551724137931	64.2312920589674\\
78.9655172413793	65.473924662951\\
73.2758620689655	66.0279905103367\\
67.5862068965517	65.8747869361726\\
61.8965517241379	65.1374779239881\\
56.2068965517241	62.9674731857184\\
50.5172413793103	59.9009479040219\\
44.8275862068966	55.6359890717561\\
39.1379310344828	48.6659489228228\\
33.448275862069	38.6054894353242\\
27.7586206896552	23.2683337816235\\
22.0689655172414	0\\
16.3793103448276	0\\
10.6896551724138	0\\
5	0\\
}--cycle;

\addplot [color=black,solid,line width=2.0pt,forget plot] 
  table[row sep=crcr]{%
5	6.45568651835702\\
10.6896551724138	10.3265654712407\\
16.3793103448276	15.1873519936888\\
22.0689655172414	20.2794236597704\\
27.7586206896552	25.6528944060518\\
33.448275862069	31.3214139100747\\
39.1379310344828	37.2074052003763\\
44.8275862068966	43.3787304398405\\
50.5172413793103	49.6507805431484\\
56.2068965517241	56.0003309825131\\
61.8965517241379	62.3898719162942\\
67.5862068965517	68.6175585217403\\
73.2758620689655	74.6285452694065\\
78.9655172413793	80.2893391555062\\
84.6551724137931	85.4698289776398\\
90.3448275862069	90.1034910480368\\
96.0344827586207	93.9566524996283\\
101.724137931034	96.9693976750291\\
107.413793103448	99.0336867308899\\
113.103448275862	100\\
118.793103448276	99.7946853786165\\
124.48275862069	98.3725284660998\\
130.172413793103	95.6135141714207\\
135.862068965517	91.4793445909217\\
141.551724137931	85.838303797511\\
147.241379310345	78.5576349402054\\
152.931034482759	69.5343748246832\\
158.620689655172	58.8521769403945\\
164.310344827586	46.8088571718135\\
170	33.9066170835904\\
};

\addplot [color=black,dash pattern=on 4pt off 2pt,line width=2.0pt,forget plot] 
  table[row sep=crcr]{%
5	4.07660735177935\\
10.6896551724138	8.80755193653309\\
16.3793103448276	13.6956651564215\\
22.0689655172414	18.8111128238237\\
27.7586206896552	24.2132117422408\\
33.448275862069	29.9152541546209\\
39.1379310344828	35.7997335072582\\
44.8275862068966	41.9698917675947\\
50.5172413793103	48.2712265602216\\
56.2068965517241	54.6105280787583\\
61.8965517241379	60.9226473408588\\
67.5862068965517	67.1812595870442\\
73.2758620689655	73.1483473347269\\
78.9655172413793	78.7567736002911\\
84.6551724137931	83.8809909236942\\
90.3448275862069	88.4321946250314\\
96.0344827586207	92.2004382669321\\
101.724137931034	95.1125106069449\\
107.413793103448	97.077410515815\\
113.103448275862	97.9173832018995\\
118.793103448276	97.5850028498196\\
124.48275862069	96.0159176577716\\
130.172413793103	93.0861179182126\\
135.862068965517	88.7741855824484\\
141.551724137931	82.9125177066529\\
147.241379310345	75.3923237287498\\
152.931034482759	66.1246736579946\\
158.620689655172	55.190729351679\\
164.310344827586	42.8534799464166\\
170	29.6726198856416\\
};

\addplot [color=mycolor1,dash pattern=on 4pt off 2pt,line width=2.0pt,forget plot] 
  table[row sep=crcr]{%
5	48.3663412572705\\
10.6896551724138	49.9465508249897\\
16.3793103448276	50.0000013524155\\
22.0689655172414	49.7106078839677\\
27.7586206896552	60.871986226711\\
33.448275862069	67.9243033038532\\
39.1379310344828	72.2092478093718\\
44.8275862068966	74.8268536018229\\
50.5172413793103	75.9740125186291\\
56.2068965517241	76.407244087759\\
61.8965517241379	76.2804501784168\\
67.5862068965517	75.328355076074\\
73.2758620689655	73.9781860784896\\
78.9655172413793	72.0635005601362\\
84.6551724137931	69.9209754834707\\
90.3448275862069	67.3025819837616\\
96.0344827586207	64.2718919808852\\
101.724137931034	61.1854090607443\\
107.413793103448	57.8088320916868\\
113.103448275862	53.9979985545881\\
118.793103448276	50.0766537774161\\
124.48275862069	45.9800425388596\\
130.172413793103	41.8236965875606\\
135.862068965517	37.3851513281464\\
141.551724137931	32.8123951482567\\
147.241379310345	28.2425921273041\\
152.931034482759	23.4879150115317\\
158.620689655172	18.5876958697509\\
164.310344827586	13.7127945834205\\
170	8.75138669085144\\
};
\addplot [color=mycolor1,solid,line width=2.0pt,forget plot] 
  table[row sep=crcr]{%
5	96.7326798980855\\
10.6896551724138	99.89309894804\\
16.3793103448276	100\\
22.0689655172414	99.4212130787597\\
27.7586206896552	98.475636008199\\
33.448275862069	97.2431145421204\\
39.1379310344828	95.7525441059764\\
44.8275862068966	94.0177155888695\\
50.5172413793103	92.0470746435184\\
56.2068965517241	89.8470125595897\\
61.8965517241379	87.4234200681899\\
67.5862068965517	84.7819209227678\\
73.2758620689655	81.9283794306184\\
78.9655172413793	78.6530743298887\\
84.6551724137931	75.6106568628334\\
90.3448275862069	72.1603034105438\\
96.0344827586207	68.42203714233\\
101.724137931034	64.7151701284196\\
107.413793103448	60.7374531465603\\
113.103448275862	56.3912326898868\\
118.793103448276	52.2606760580513\\
124.48275862069	47.661689061767\\
130.172413793103	43.192983936458\\
135.862068965517	38.6717208888445\\
141.551724137931	33.7611393800196\\
147.241379310345	29.0162707737045\\
152.931034482759	24.1109871186643\\
158.620689655172	19.0900890706585\\
164.310344827586	14.010533728164\\
170	8.95773755843823\\
};
\end{axis}
\end{tikzpicture}%
    \end{subfigure}
    \caption{Normalized step distance $S/\max\{S\}$ versus phase difference -- nominal-friction distance $S$ (solid curves), \colorbox{black!30}{$\mu_1/\mu_2 \in [0.9 , 1.1]$ (shaded areas)} and uncertain-friction distance $S_\mu$ (dashed curves). \textcolor{myblue}{$\omega=0.3$ [rad/s] in blue} and $\omega=16$ [rad/s] in black}
    \label{fig:S_phi_dmu}
\end{figure}
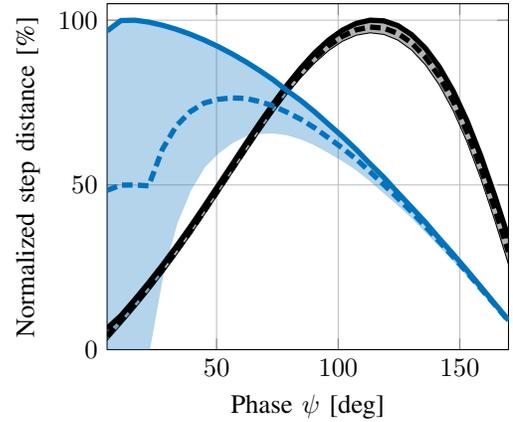

\begin{figure}
    \centering
    \begin{subfigure}{0.9\linewidth}
	    \input{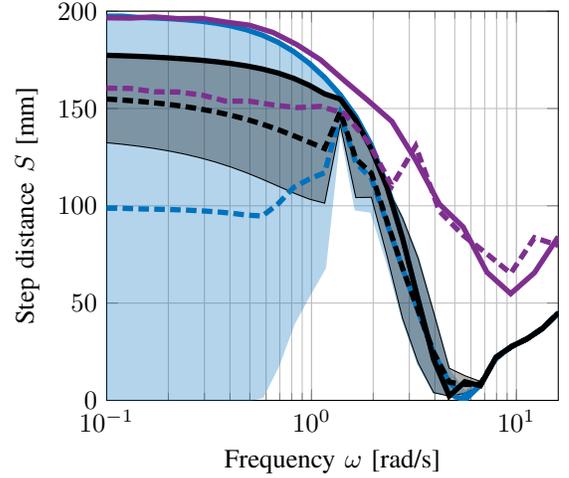}
    \end{subfigure}
    \caption{Step distance versus frequency -- nominal-friction distance $S$ (solid curves), \colorbox{black!30}{$\mu_1/\mu_2 \in [0.9 , 1.1]$ (shaded areas)} and uncertain-friction distance $S_\mu$ (dashed curves). \textcolor{myblue}{Optimization ($\psi^*(\omega)$) of the nominal-friction distance $S$ in blue} and optimization ($\psi^*_\mu(\omega)$) of the uncertain-friction distance $S_\mu$ in black. \textcolor{mypurple}{Optimization of the nominal-friction distance $S$ (solid purple curve) and $S_\mu$ (dashed purple curve) with inputs' time scaling}}
    \label{fig:S_w_opt}
\end{figure}

Based on the results from our experiments and previous works \cite{gamus2020understanding,majidi2013influence}, practical considerations suggest that the simulation predictions must be tested for sensitivity to inaccuracies in friction. Modeling (or measuring) and simulating the actual distribution of the friction along various surfaces involves increased complexity. In this section we propose a deterministic and computationally-cheap criterion, which captures the effects of uneven friction, and allows for numerical optimization. We perform the sensitivity analysis by varying the friction coefficients at each of the two contacts such that $\mu_i \in [1-\varepsilon,1+\varepsilon] \mu$ (where $\mu_i$ is the coefficient at contact $i$ and $\mu$ is the nominal measured coefficient). A symmetric perturbation $\mu_1=\mu_2=(1\pm\varepsilon)\mu$ was found to insignificantly affect the distance $S$ for reasonable $\varepsilon$ (not shown). However, for an asymmetric perturbation, Fig.\,\ref{fig:freq} shows in the shaded blue area how the distance $S$ decreases as $\mu_1/\mu_2 \rightarrow 1 \pm \varepsilon$ for $\varepsilon=0.1$ (which is close to the actual range measured in calibration experiments). Assuming the friction varies along the gait and among the strides within this uncertainty range (i.e. within the shaded area), we propose the following empirical computationally-simple criterion of robustness to uncertainties in friction
\begin{equation}\label{eq:Savg}
    S_{\mu}=\frac{1}{2}\left(S_\text{min}+S_\text{max}\right),
\end{equation}
where
\begin{equation}
    S_\text{min/max}=\min/\max \left\{ 
    S_{\mu_i \in [1-\varepsilon,1+\varepsilon] \mu} \right\}.
\end{equation}
This \textit{uncertain-friction distance} is in fact the mean of the possible distances in the shaded area, as depicted by the dashed blue curve in Fig.\,\ref{fig:freq}(a). We observe that $S_\mu$ falls with excellent agreement to the experimental results, proving the validity of the proposed criterion. Furthermore, the width of the experiments' error bars, that represent the variance of the measurements, agree qualitatively with the width of the shaded area -- hence, this analysis also predicts the repeatability of the gaits. Note that the following legend is consistent in all graphs analysing the step distance (i.e. Fig.\,\ref{fig:freq}(a), \ref{fig:S_phi_dmu} and \ref{fig:S_w_opt}): the solid curve depicts the \textit{nominal-friction distance} $S$ (without friction perturbations), the shaded area (of similar color) shows the range of $S$ for friction perturbation $\mu_1/\mu_2 \rightarrow 1 \pm \varepsilon$ and the dashed curve (of similar color) depicts $S_\mu$. A solid curve connecting error bars (if exists) shows the experimental results in the corresponding parameters.

We deduce that the low frequency range is highly sensitive to friction uncertainties, while increasing the actuation frequency allows for gaits with similar (and even higher) average velocity that show smaller sensitivity, thus having increased robustness to variations in surface friction. Further analyzing the robustness via the proposed criterion $S_\mu$, shows that significant improvement can be achieved at low frequencies by optimizing the phase difference. This is illustrated in Fig.\,\ref{fig:S_phi_dmu} for $\omega=0.3$ [rad/s] (in blue) and $\omega=16$ [rad/s] (in black), normalized by the maximal $S$ (since the dimensional $S$ for $\omega=16$ [rad/s] is much smaller). We see how at low frequencies choosing the phase $\psi=16^{\circ}$ at the optimum of $S$ (solid curve), results in a decrease of 50\% in the presence of friction uncertainties, as indicated by $S_\mu$ (dashed curve). On the other hand, the optimum of $S_\mu$ at $\psi=56^{\circ}$ gives only 24\% decrease of the step distance, with much narrower shaded area -- which shall result in less variance and more robust gait. As the frequency rises (black curves), $S$ and $S_\mu$ become closer, as previously deduced. This is also illustrated by plotting $\psi^*_\mu(\omega)$, the optimal phase to maximize the uncertain-friction distance $S_\mu$, in dashed blue curve over the $\psi-\omega$ map in Fig.\,\ref{fig:psi_w_map}. Note how this curve merges with $\psi^*(\omega)$, the optimal phase to maximize the nominal-friction distance $S$ (dotted black curve), at high frequencies, which we have seen to be naturally robust from the dynamics of the system.

In Fig.\,\ref{fig:S_w_opt} we study the gaits' performance under uncertainties in friction for various frequencies, at phase $\psi^*(\omega)$ (in blue) versus at phase $\psi^*_\mu(\omega)$ (in black). Though the nominal-friction distance $S$ (solid curves), which is the optimistic prediction for perfectly even friction, is slightly lower for $\psi=\psi^*_\mu$, the uncertain-friction distance $S_\mu$ (dashed curves), which was shown to be a realistic prediction, is significantly improved at low frequencies. The experimental results for $\psi=\psi^*_\mu$ are plotted in Fig.\,\ref{fig:freq}(a) by a black curve with error bars. We see significant improvement of the distance and repeatability, compared to unoptimized gaits for $\psi=20^{\circ}$ in purple curve. Comparing these observations to the analytical prediction in gray curves (which are the same as the black curves in Fig.\,\ref{fig:S_w_opt}) we see excellent agreement at high frequencies. At low frequencies the improvement is smaller than predicted by the model and our uncertain-friction criterion.


\section{Mass asymmetry and time-scaling input shaping}\label{sec:learning}

In this section we investigate the influence of mass distribution on the step distance $S$ and discover the effects of asymmetry. We introduce a time-scaling input shaping strategy, which achieves asymmetry without mechanical changes to the robot, and utilize machine-learning algorithm to optimize the parameters of such inputs.

Considering the ratio between the mass of the distal and center links $\frac{1}{2}(m_1+m_2)/m_0$, the distance $S$ grows almost monotonically with the decrease of the distal links' mass (see Appendix). A more divergent and influential parameter is the asymmetry of mass distribution among the distal links $m_1/m_2$ (also shown in Supplemental material). Hereafter we also keep fixed amplitude $A=18^{\circ}$ and nominal angle $\varphi_0=110^{\circ}$, in order to investigate only the effect of asymmetry. At the low frequencies range ($\omega\lesssim 1$ [rad/s]), the gaits are highly sensitive to asymmetry in any direction, causing significant decrease in $S$. As the frequency increases, the sensitivity is reduced and asymmetric mass distribution even becomes advantageous. The preference moves from heavier front link $m_1>m_2$ at mid-range frequencies (where even direction reversal occurs) to heavier rear link $m_2>m_1$ at high frequencies. Exploiting the advantages of asymmetric mass distribution requires mechanically reassembling the robot for each frequency. Therefore, we desire to achieve a similar improvement effect of the distance $S$ by shaping the input signals $\varphi_i(t)$.

\begin{figure}
    \centering
    \begin{subfigure}{0.65\linewidth}
%
%
\definecolor{mycolor1}{rgb}{0.00000,0.44700,0.74100}%
\definecolor{mycolor2}{rgb}{0.49400,0.18400,0.55600}%
\begin{tikzpicture}

\begin{axis}[%
extra description/.code={\node[anchor=north east] at ([xshift=-20pt]current axis.north west) {(a)};},
width=0.75\textwidth,
at={(0.758in,0.567in)},
scale only axis,
xmin=7.80519913935352,
xmax=9.7564989241919,
xtick={7.80519913935352,8.29302408556312,8.78084903177271,9.26867397798231,9.7564989241919},
xticklabels={{0},{$\frac{\pi}{2}$},{$\pi$},{$\frac{3\pi}{2}$},{$2\pi$}},
xlabel={$\omega t=3.2 t$},
xmajorgrids,
ymin=90,
ymax=130,
ylabel={$\varphi_i$ [deg]},
ymajorgrids,
axis background/.style={fill=white},
legend style={legend cell align=left,align=left,draw=white!15!black}
]

\addplot [color=black,dash pattern=on 4pt off 2pt,line width=2.0pt,forget plot]
  table[row sep=crcr]{%
7.8065063665819	121.653166702017\\
7.86698019852072	118.776795947786\\
7.93051482701415	115.39927362807\\
7.99517818009004	111.731185539846\\
8.05710549210098	108.145982506866\\
8.11102438189119	105.07936575142\\
8.16024942519498	102.406770390618\\
8.20656577252199	100.064961216657\\
8.25112051680866	98.0200010874751\\
8.29482582173957	96.25320178639\\
8.33676221126202	94.8133135527584\\
8.36291177843326	94.0541117227498\\
8.40803957404415	93.0125783948207\\
8.45065416232896	92.3578201805902\\
8.49176841962059	92.040710087195\\
8.53244522166507	92.0365975098729\\
8.57410578442408	92.3516705150436\\
8.61639627436247	92.9961353404442\\
8.65787398704445	93.9344445455465\\
8.69945249851849	95.1625085663304\\
8.7381313808079	96.5441360503951\\
8.7718026270089	97.9176206202663\\
8.80134000631978	99.2400185556923\\
8.8375007370076	100.990213266232\\
8.87527409610154	102.948495537765\\
8.91621815686987	105.188000440746\\
8.95684732705872	107.493028298787\\
8.99719087268137	109.824384762711\\
9.03838923057251	112.207908471762\\
9.07908291116173	114.524272440408\\
9.1197094315198	116.759376377599\\
9.16146419231954	118.935704994866\\
9.20550484909299	121.055488089855\\
9.24951509052246	122.951858197706\\
9.29204245576744	124.537746529407\\
9.33438358591984	125.846044770011\\
9.37739232524432	126.873464039057\\
9.42181806244377	127.594965376377\\
9.46843617803927	127.965020957091\\
9.5181943018875	127.913717222187\\
9.57218161667164	127.338809699441\\
9.6224777526277	126.331844027401\\
9.64299526142134	125.796307255136\\
9.67451552538006	124.840077605457\\
9.70603578933879	123.730948192272\\
9.73755605329752	122.480346558583\\
9.75141813885848	121.888831517121\\
};

\addplot [color=mycolor1,dash pattern=on 4pt off 2pt,line width=2.0pt,forget plot]
  table[row sep=crcr]{%
7.8065063665819	122.120935584852\\
7.86698019852072	124.467865143485\\
7.93051482701415	126.342553483824\\
7.99517818009004	127.549588563358\\
8.05710549210098	127.994523557799\\
8.11102438189119	127.80045082532\\
8.16024942519498	127.154979590511\\
8.20656577252199	126.154079890143\\
8.25112051680866	124.852120954765\\
8.29482582173957	123.277981389258\\
8.33676221126202	121.520034970254\\
8.36291177843326	120.315380794946\\
8.40803957404415	118.069488286581\\
8.45065416232896	115.791507528148\\
8.49176841962059	113.489929848608\\
8.53244522166507	111.152528622769\\
8.57410578442408	108.738448368542\\
8.61639627436247	106.31129792607\\
8.65787398704445	103.996929283944\\
8.69945249851849	101.784393028875\\
8.7381313808079	99.8575217259208\\
8.7718026270089	98.3071859599607\\
8.80134000631978	97.0597926094879\\
8.8375007370076	95.6931695353216\\
8.87527409610154	94.4729806937871\\
8.91621815686987	93.4102445152111\\
8.95684732705872	92.6405687988922\\
8.99719087268137	92.1701569167087\\
9.03838923057251	92.0002144158352\\
9.07908291116173	92.1431952618225\\
9.1197094315198	92.5916636719982\\
9.16146419231954	93.3627922948034\\
9.20550484909299	94.5015232911227\\
9.24951509052246	95.9506329911091\\
9.29204245576744	97.6191504631754\\
9.33438358591984	99.5109343252344\\
9.37739232524432	101.631848813704\\
9.42181806244377	103.990560547968\\
9.46843617803927	106.596922641604\\
9.5181943018875	109.46188007972\\
9.57218161667164	112.583580806598\\
9.6224777526277	115.423605046163\\
9.64299526142134	116.545453379467\\
9.67451552538006	118.21154148987\\
9.70603578933879	119.793024735987\\
9.73755605329752	121.273608835485\\
9.75141813885848	121.888831517121\\
};

\addplot [color=mycolor1,solid,line width=2.0pt,forget plot]
  table[row sep=crcr]{%
7.80537264118227	113.269452712035\\
7.83982874848512	116.238916378548\\
7.87181566826618	118.912858682245\\
7.89500983901747	120.744393771419\\
7.90481885205727	121.481011279607\\
7.93622614686382	123.645846805387\\
7.96545486559	125.336222468771\\
7.9919442885953	126.543889861032\\
8.01718786268902	127.371863601076\\
8.04196922023438	127.855037803323\\
8.06686039949901	127.998771415025\\
8.09242303891478	127.78745003689\\
8.11944190050721	127.17830280055\\
8.14927423591679	126.07324345989\\
8.17979633862364	124.522217669354\\
8.20980892736486	122.647263961507\\
8.24011102071553	120.478293947009\\
8.27150443202474	118.02581317887\\
8.30511535421165	115.270536762421\\
8.3419022823094	112.222503633696\\
8.37924801228166	109.21285871285\\
8.41250520576852	106.685524248655\\
8.44269080444244	104.561787485528\\
8.47151183660566	102.70960262796\\
8.49982238923	101.069391073021\\
8.52823120316589	99.6058736406572\\
8.55731345201002	98.2940949858422\\
8.58771059694878	97.1154483607726\\
8.62031218771696	96.0536780010348\\
8.65657132782448	95.0928160384712\\
8.69937988702992	94.2135623157695\\
8.7455110471729	93.5170716517824\\
8.78714236840993	93.0621530507717\\
8.8181440177947	92.8059990388193\\
8.86145228309672	92.5379570740011\\
8.90567674862404	92.3445256857013\\
8.94389240078774	92.2243067824607\\
8.97641481781984	92.147344509454\\
9.00631379224193	92.0927737035841\\
9.03461946268588	92.0531870541243\\
9.06220806354266	92.0249459561936\\
9.08969811467606	92.006958792875\\
9.11777269909603	92.0000129309089\\
9.14758104364703	92.0075668576236\\
9.18069754150827	92.0387327432212\\
9.21806837550504	92.1123343838714\\
9.2524714398225	92.2290722089311\\
9.28384386496492	92.3903088731094\\
9.31459929684233	92.6140019253688\\
9.34429294550039	92.907404638919\\
9.36388231374269	93.1509858800129\\
9.39383278524673	93.6142882097386\\
9.42470503744416	94.2261065481493\\
9.45387421456284	94.9501335098239\\
9.48054252073252	95.7530314564922\\
9.50609174174501	96.6617924785638\\
9.53126150704095	97.7011298277646\\
9.55666507169327	98.9032549021557\\
9.5830087893949	100.31770682736\\
9.59309074951589	100.904628367781\\
9.62265374689407	102.769145362436\\
9.65378547820027	104.953366250282\\
9.68774577192567	107.564304000998\\
9.72090619631161	110.293881163565\\
9.73225871227332	111.257336738255\\
9.7377328341548	111.725558043889\\
9.74320695603627	112.195633285498\\
9.74615543645442	112.449454947268\\
9.7475403581397	112.568804491348\\
9.74892527982498	112.688226521229\\
9.75031020151026	112.807714313622\\
9.75141813885848	112.903347348845\\
};

\addplot [color=black,solid,line width=2.0pt,forget plot]
  table[row sep=crcr]{%
7.80537264118227	112.98747264052\\
7.83982874848512	110.177098787653\\
7.87181566826618	107.685052376527\\
7.89500983901747	105.979506949464\\
7.90481885205727	105.28826030255\\
7.93622614686382	103.207501948224\\
7.96545486559	101.464377131166\\
7.9919442885953	100.051355960277\\
8.01718786268902	98.8517508745547\\
8.04196922023438	97.8094706201085\\
8.06686039949901	96.8908984331655\\
8.09242303891478	96.0724348042454\\
8.11944190050721	95.3332354308427\\
8.14927423591679	94.6514612888346\\
8.17979633862364	94.0813143963481\\
8.20980892736486	93.6278952819599\\
8.24011102071553	93.2605248151805\\
8.27150443202474	92.9588785574716\\
8.30511535421165	92.7076595389388\\
8.3419022823094	92.4991753335269\\
8.37924801228166	92.3415173228709\\
8.41250520576852	92.2356076237529\\
8.44269080444244	92.1610734236498\\
8.47151183660566	92.1052420470074\\
8.49982238923	92.0626611278766\\
8.52823120316589	92.0309006587219\\
8.55731345201002	92.0095162147141\\
8.58771059694878	92.0001389491859\\
8.62031218771696	92.0075597388908\\
8.65657132782448	92.0431667679383\\
8.69937988702992	92.1364200210257\\
8.7455110471729	92.3264795604078\\
8.78714236840993	92.6122937314544\\
8.8181440177947	92.9200367675807\\
8.86145228309672	93.5265148089057\\
8.90567674862404	94.415078987111\\
8.94389240078774	95.4538903722945\\
8.97641481781984	96.5695095434583\\
9.00631379224193	97.8039015992578\\
9.03461946268588	99.1693477210898\\
9.06220806354266	100.690595154748\\
9.08969811467606	102.393280348749\\
9.11777269909603	104.317268202283\\
9.14758104364703	106.546569726591\\
9.18069754150827	109.210534312611\\
9.21806837550504	112.377864977163\\
9.2524714398225	115.348497787834\\
9.28384386496492	118.00868162759\\
9.31459929684233	120.480392991848\\
9.34429294550039	122.654618855744\\
9.36388231374269	123.936213598433\\
9.39383278524673	125.606470024608\\
9.42470503744416	126.902003953563\\
9.45387421456284	127.679618167438\\
9.48054252073252	127.985330839241\\
9.50609174174501	127.906730009166\\
9.53126150704095	127.477946082508\\
9.55666507169327	126.706632833335\\
9.5830087893949	125.574463819291\\
9.59309074951589	125.059225374574\\
9.62265374689407	123.317798931695\\
9.65378547820027	121.172525225332\\
9.68774577192567	118.56913547664\\
9.72090619631161	115.869866296856\\
9.73225871227332	114.929007433234\\
9.7377328341548	114.474356627235\\
9.74320695603627	114.019646227772\\
9.74615543645442	113.774848828786\\
9.7475403581397	113.659918806474\\
9.74892527982498	113.545031543577\\
9.75031020151026	113.430193430099\\
9.75141813885848	113.33836259171\\
};
\end{axis}
\end{tikzpicture}%
    \label{fig:phi_timescaling}
    \end{subfigure}
    ~
    \begin{subfigure}{0.65\linewidth}
	    \input{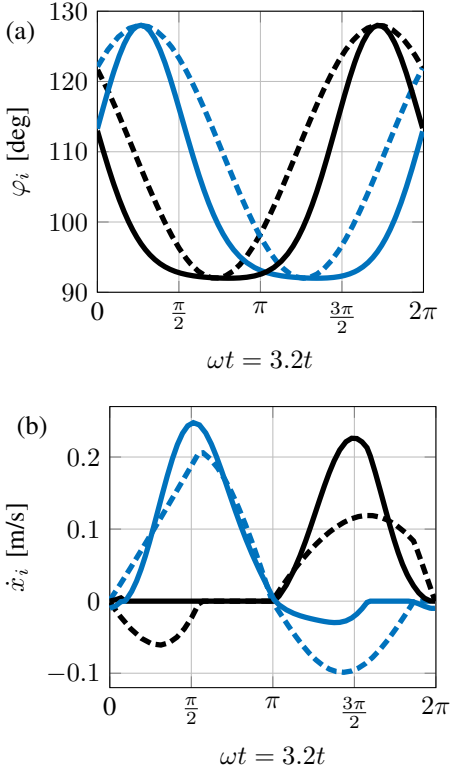}
        \label{fig:dx_timescaling}
    \end{subfigure}
    \caption{Gait with time-scaling (solid curves) and without time-scaling $\alpha=0$ (dashed curves) for $\omega=3.2$ [rad/s]. (a) Joint angles \textcolor{myblue}{$\varphi_1$ (blue)} and $\varphi_2$ (black). (b) Contacts' velocities \textcolor{myblue}{$\dot{x}_1$ (blue)} and $\dot{x}_2$ (black)}
    \label{fig:timescaling}
\end{figure}

We propose a \textit{time-scaling} strategy, where the joint angles remain of the form $\varphi_i(t)~=~\varphi_0~+~A~\sin \big( s(\omega~t~\pm~\psi/2) \big)$ as in (\ref{eq:ref}), but a time-scaling function $s(\xi)$ is introduced as
\begin{equation}\label{eq:timescaling}
    s(\xi)=\xi+\alpha \sin (\xi+\Upsilon).
\end{equation}
For $\alpha=0$ these functions are reduced to the unscaled case of (\ref{eq:ref}), but $\alpha\neq0$ generates an uneven ``internal'' pace in $\varphi_i(s(t))$, while preserving the overall shape and the principal harmonic (as long as $\dot{s}(t)\geq0$). This is illustrated in Fig.\,\ref{fig:timescaling}(a) for optimal $\psi=\psi^*$ without time-scaling $\alpha=0$ (dashed curves), compared to optimal choice of both $(\alpha,\Upsilon)$ to maximize $S$ and same $\psi^*$ (solid curves), at $\omega=3.2$ [rad/s]. Fig.\,\ref{fig:timescaling}(b) shows how these time-scaled inputs significantly improve $S$ by both increasing $\dot{x}_i(t)$ in the advancement direction and reducing slippage in the opposing direction. At this frequency, the optimal scaled input achieves this improvement by extending the legs rapidly and retracting them slowly  (see supplementary video \cite{suppVid}).

Since the system is highly nonlinear and non-smooth, induced by the hybrid contact-state transitions, finding optimal values of multiple parameters involves increased complexity. Gradient-based optimization algorithms (like Newton's method or gradient descent) struggle and typically converge to a local minimum, which depends on the basin of attraction the starting point \cite{calandra2014experimental}. To search for global optimum, optimizations must be initialized from multiple points, which is time-consuming (we attempted to use MATLAB\textregistered\ \texttt{fmincon} with \texttt{multistart}). Contrarily, \textit{Bayesian optimization} is a \textit{machine-learning} approach which is based on fitting a Gaussian process \cite{rasmussen2003gaussian}. This algorithm was shown to be very efficient for legged robotic locomotion \cite{calandra2014experimental, tesch2011using}, probably since such systems exhibit non-smooth behavior and coupling of parameters. We applied this machine-learning algorithm (via MATLAB\textregistered\ \texttt{bayesopt}) to our numerical simulation, described in Section \ref{sec:analysis}, and achieved rapid and stable convergence (compared to gradient-based methods). Fig.\,\ref{fig:S_w_opt} illustrates the machine-learning optimization of time-scaling parameters and phase difference $(\alpha,\Upsilon,\psi)$ to maximize $S$ (solid purple curve) and $S_\mu$ (dashed purple curve) for various frequencies. A significant improvement of $S$ is achieved at the middle- and high-frequency ranges.


\section{Hybrid feedback control}\label{sec:control}
In the approach presented so far, we prescribed harmonic input signals to the actuated joints and optimize their parameters, assuming minimal sensing and computation abilities (just the closed-loop control of the servos). This often leads to undesired slippage, which reduces the distance the robot travels. An alternative approach is to prescribe a feedback control law, which utilizes sensing of the contacts' velocities to eliminate undesired slippage. Most existing works on slippage in robotic locomotion, either slippage detection and recovery \cite{takemura2005slip} or avoidance (for quasistatic locomotion) \cite{samadi2020icra}, deal with systems with higher dimensionality, which allows to stabilize the robots' main body separately from the contacts. The inchworm crawling locomotion has inherent continuous slippage, and the three-link robot has very few DoF and hence requires a different method.

We propose a hybrid (contacts state-based) feedback control law, based on PD reference tracking with inverse dynamics \cite{murray1994mathematical}, assuming the whole state $\mathbf{q},\dot{\mathbf{q}}$ can be measured (including contacts' velocities $\dot{x}t_i$). Once slip-slip state is detected (indicating undesired slippage), the hybrid control switches from tracking of a reference trajectory $\mathbf{q}_\text{c}(t)$ in (\ref{eq:ref}) to tracking of ``ideal'' contacts velocities. The detailed derivation is given in the Appendix. 
Remarkably, simulation results of this control law show drastic improvement of travelled distance by suppressing undesired slippage (see Appendix for examples). Nevertheless, the results require careful tuneup and currently work only for a range of reference input frequencies. We deduce that for high frequencies the three-link robot's slippage is strongly enforced by the natural dynamics of the contact and can hardly be eliminated. For most parameter values the control struggles to overcome slippage, which results in either loss of contact or worse performance than without feedback. These limitation can perhaps be solved by forces sensing-based control \cite{kojima2015shuffle} and additional actuation.

\section{Concluding discussion}
This paper has addressed modeling of dynamic robotic inchworm-like crawling with passive frictional contacts. Such locomotion is sensitive to uncertainties in the friction, but once that is accounted for (with a minimal analysis we propose) the model shows good agreement with experimental results.

The model also allows for comprehension of the influence of inertia and other phenomena. For example it is observed that with low-frequency actuation, tuning the gait's parameters according to the criterion we proposed can significantly improve the robustness with respect to variations and unevenness in surface friction. At high frequencies the gaits demonstrate natural robustness with even higher average velocities.

The quasistatic low-frequency range was also observed to be very sensitive to asymmetry. However, as the frequency rises asymmetry can actually be exploited to improve the gaits' performance in traveling distance. Similar effect was also created by shaping the inputs with time-scaling technique, without mechanical changes to the robot. A feedback control, based on the contacts state, was also proposed. Though this approach can reduce undesired slippage for some cases, in general it struggles for the three-link configuration, and requires more advanced sensing and computation. Future research may address this with additional DoF and actuation or sensing of the contact forces.

The latter analyses were only investigated in simulation. Experimentally testing the optimizations proposed in Sections \ref{sec:fric} and \ref{sec:learning} and the control in Section \ref{sec:control} on this and other legged robots shall also be the subject of future research. 

The presented notions can be implemented in future into soft crawling robots, both quasistatic and dynamic. We also believe that crawling locomotion in articulated legged robots is advantageous and can be exploited more extensively. Finally, just as McGeer's biped \cite{mcgeer1990passive} was an inspiration to studies of simple and energetically efficient control, we hope that our three-link model will pave the way for future analytic and experimental works in inertial crawling locomotion.

\bibliography{mybibfile}


\clearpage
\thispagestyle{empty}

\begin{strip}
\begin{equation}\label{eq:M}
    \mathbf{M}=
    \begin{bmatrix}
    m_0 +2m & 0 & \frac{l m}{2}(\text{s}_1+\text{s}_2) & -\frac{l m}{2}\text{s}_1 & \frac{l m}{2} \text{s}_2\\
    0 & m_0 +2m & \frac{l m}{2} (\text{c}_1-\text{c}_2) & -\frac{l m}{2} \text{c}_1 & -\frac{l m}{2} \text{c}_2\\
    \frac{l m}{2} (\text{s}_1+\text{s}_2) & \frac{l m}{2} (\text{c}_1-\text{c}_2) & \text{M}_{3,3} & \frac{l m}{4}(l_0\cos\varphi_1-l)-J & J+\frac{l m}{4}(l-l_0\cos\varphi_2)\\
    -\frac{l m}{2}\text{s}_1 & -\frac{l m}{2}\text{c}_1 & \frac{l m}{4}(l_0\cos\varphi_1-l)-J & \frac{l^2 m}{4}+J & 0\\
    \frac{l m}{2} \text{s}_2 & -\frac{l m}{2} \text{c}_2 & J+\frac{l m}{4}(l-l_0\cos\varphi_2) & 0 & \frac{l^2 m}{4}+J
    \end{bmatrix}
\end{equation}
\end{strip}

\section*{Appendix}\label{sec:appendix}
\subsection*{Motion equations}
The terms for the matrices in the dynamic motion equations (\ref{eq:eom}) are as follows: the mass matrix $\mathbf{M}$ is given in (\ref{eq:M}), where $\text{M}_{3,3}=2J+J_{0}+\frac{m}{2} \big({l_0}^2+l^2-l_0 l \cos\varphi_1 -l_0 l \cos\varphi_2 \big)$ and the abbreviations $\text{s}_1\equiv\sin(\varphi_1-\theta)$, $\text{s}_2\equiv\sin(\varphi_2+\theta)$, $\text{c}_1\equiv\cos(\varphi_1-\theta)$ and $\text{c}_2\equiv\cos(\varphi_2+\theta)$. Also
\begin{equation}
    \mathbf{B}=\frac{1}{2} l m
    \begin{bmatrix}
    \left(\dot\varphi_2+\dot\theta\right)^2\text{c}_2-\left(\dot\varphi_1-\dot\theta \right)^2 \text{c}_1\\
    \left(\dot\varphi_2+\dot\theta\right)^2\text{s}_2-\left(\dot\varphi_1+\dot\theta \right)^2 \text{s}_1\\
    \text{B}_3\\
    -\frac{1}{2} l_0 {\dot{\theta }}^2 \sin\varphi_1\\
    -\frac{1}{2} l_0 {\dot{\theta }}^2 \sin\varphi_2
    \end{bmatrix},
\end{equation}
where $\text{B}_3=\frac{1}{2} l_0 \Big(\big(2\dot\theta \dot\varphi_1-{\dot\varphi_1}^2\big)\sin\varphi_1+\Big({\dot\varphi_2}^2+2\dot\theta \dot\varphi_2 \big) \sin\varphi_2\Big)$, and
\begin{equation}
    \mathbf{G}=g
    \begin{bmatrix}
    0\\
    m_0+2m\\
    \frac{1}{2} l m (\text{c}_1-\text{c}_2)\\
    -\frac{1}{2} l m \text{c}_1\\
    -\frac{1}{2} l m \text{c}_2
    \end{bmatrix}.
\end{equation}
The Jacobians in (\ref{eq:Jac}) are
\begin{equation}
    \mathbf{W}_1
    \equiv
    \begin{bmatrix}\mathbf{w}_1^x\\\mathbf{w}_1^y\end{bmatrix}
    =
    \begin{bmatrix}
    1 & 0 & \frac{1}{2} (2l\text{s}_1+l_0\sin\theta ) & -l\text{s}_1 & 0\\
    0 & 1 & \frac{1}{2} (2l\text{c}_1-l_0\cos\theta) & -l\text{c}_1 & 0
    \end{bmatrix}
\end{equation}
and
\begin{multline}
    \mathbf{W}_2
    \equiv
    \begin{bmatrix}\mathbf{w}_2^x\\\mathbf{w}_2^y\end{bmatrix}=
    \\=
    \begin{bmatrix}
    1 & 0 & \frac{1}{2}(2l\text{s}_2-l_0\sin\theta ) & 0 & l\text{s}_2\\
    0 & 1 & -\frac{1}{2}(2l\text{c}_2-l_0\cos\theta) & 0 & -l\text{c}_2
    \end{bmatrix}.
\end{multline}

When one of the contacts is at slippage while the other maintains contact sticking -- for concreteness let us consider stick-slip contact-state -- from (\ref{eq:v}) we get three kinematic constraints: contact-sticking of one leg $\mathbf{v}_1=\mathbf{W}_1\dot{\mathbf{q}}=0$ and no-detachment of the slipping leg $\dot{y}_2=\mathbf{w}_2^y\dot{\mathbf{q}}=0$. Differentiating these constraints with respect to time gives
\begin{equation}\label{eq:const_der}
    \begin{bmatrix}\mathbf{W}_1\\ \mathbf{w}_2^y\end{bmatrix}\ddot{\mathbf{q}}=-\begin{bmatrix}\dot{\mathbf{W}}_1\\ \dot{\mathbf{w}}_2^y\end{bmatrix}\dot{\mathbf{q}}.
\end{equation}
We can now formulate the dynamic equations (\ref{eq:eom}) with the constraints (\ref{eq:const_der}) as
\begin{equation}\label{eq:full}
    \begin{bmatrix}\mathbf{M} & -\mathbf{W}_1\!^\text{T} & -\mathbf{W}_2\!^\text{T}\boldsymbol{\Gamma}_2\\
    \mathbf{W}_1 & 0\quad 0 & 0\\
    \mathbf{w}_2^y & 0\quad 0 & 0\\
    \end{bmatrix}
    \begin{bmatrix}\ddot{\mathbf{q}}\\ f_1^x \\ f_1^y \\ f_2^y\end{bmatrix}=
    \begin{bmatrix}-\mathbf{B}-\mathbf{G}+\mathbf{E}\boldsymbol{\tau}\\
    -\dot{\mathbf{W}}_1\dot{\mathbf{q}}\\ -\dot{\mathbf{w}}_2^y\dot{\mathbf{q}}
    \end{bmatrix},
\end{equation}
where 
\begin{equation}\label{eq:lam_slip}
    \boldsymbol{\Gamma}_i\equiv \begin{bmatrix}-\mu\,\text{sign}\, \dot{x}_i\\ 0\end{bmatrix},
\end{equation}
for the contact at slippage from (\ref{eq:slip}).

Slip-slip contact state is formulated similarly to (\ref{eq:full}), though the contact forces of both contacts maintain (\ref{eq:lam_slip}) and the constraints are reduced to no-detachment only $\dot{y}_1=\dot{y}_2=0$, giving
\begin{equation}\label{eq:slipslip}
    \begin{bmatrix}\mathbf{M} & -\mathbf{W}_1\!^\text{T}\boldsymbol{\Gamma}_1 & -\mathbf{W}_2\!^\text{T}\boldsymbol{\Gamma}_2\\
    \mathbf{w}_1^y & 0 & 0\\
    \mathbf{w}_2^y & 0 & 0\\
    \end{bmatrix}
    \begin{bmatrix}\ddot{\mathbf{q}}\\ f_1^y \\ f_2^y\end{bmatrix}=
    \begin{bmatrix}-\mathbf{B}-\mathbf{G}+\mathbf{E}\boldsymbol{\tau}\\
    -\dot{\mathbf{w}}_1^y\dot{\mathbf{q}}\\ -\dot{\mathbf{w}}_2^y\dot{\mathbf{q}}
    \end{bmatrix}.
\end{equation}

The mass and Jacobian matrices can decomposed into blocks corresponding to the constrained- and body-coordinates
as $\mathbf{M}=\big[\mathbf{M}_\text{c}\ \mathbf{M}_\text{b}\big]$ and $\mathbf{W}_i=\big[\{\mathbf{W}_i\}_\text{c}\ \{\mathbf{W}_i\}_\text{b}\big]$. Then we can rearrange (\ref{eq:full}) and (\ref{eq:slipslip}) such that all the unknowns are on one side as (only the formulation of stick-slip in (\ref{eq:full}) shown for brevity)
\begin{multline}\label{eq:full_b}
    \begin{bmatrix}\mathbf{M}_\text{b} & -\mathbf{E} & -\mathbf{W}_1\!^\text{T} & -\mathbf{W}_2\!^\text{T}\boldsymbol{\Gamma}_2\\
    \{\mathbf{W}_1\}_\text{b} & 0\quad 0 & 0\quad 0 & 0\\
    \{\mathbf{w}_2^y\}_\text{b} & 0\quad 0 & 0\quad 0 & 0 \\
    \end{bmatrix}
    \begin{bmatrix}\ddot{\mathbf{q}}_\text{b} \\ \boldsymbol{\tau} \\ f_1^x \\ f_1^y \\ f_2^y\end{bmatrix}=
    \\
    =\begin{bmatrix}-\mathbf{M}_\text{c}\\
    -\{\mathbf{W}_1\}_\text{c}\\
    -\{\mathbf{w}_2^y\}_\text{c}\end{bmatrix}\ddot{\mathbf{q}}_\text{c}+
    \begin{bmatrix}-\mathbf{B}-\mathbf{G}\\
    -\dot{\mathbf{W}}_1\dot{\mathbf{q}}\\ -\dot{\mathbf{w}}_2^y\dot{\mathbf{q}}
    \end{bmatrix}.
\end{multline}

Finally, for a state-vector $\mathbf{x}(t)=[{\mathbf{q}_\text{b}}^\text{T} \quad {\dot{\mathbf{q}}_\text{b}}^\text{T}]^\text{T}$ and prescribed $\mathbf{q}_\text{c}(t), \dot{\mathbf{q}}_\text{c}(t), \ddot{\mathbf{q}}_\text{c}(t)$, we get a series of complete first order ODEs.

\subsection*{Mass distribution colormaps}
\begin{figure*}
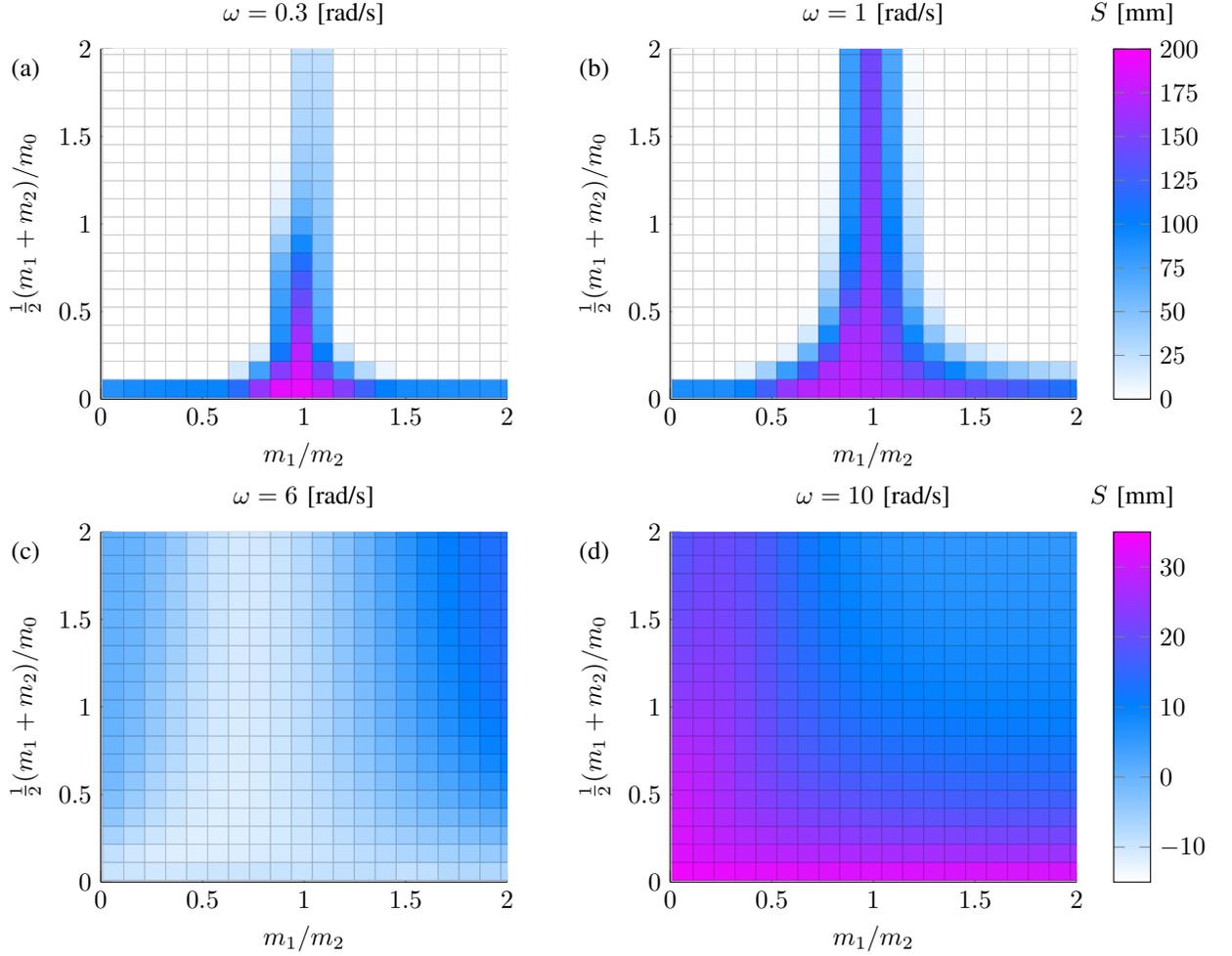

    \centering
    \begin{subfigure}{0.4\linewidth}
	    \input{supp/mass_map_w03.tikz}
    \end{subfigure}
    ~
    \begin{subfigure}{0.4\linewidth}
	    \input{supp/mass_map_w1.tikz}
    \end{subfigure}
    \\
    \centering
    \begin{subfigure}{0.4\linewidth}
	    \input{supp/mass_map_w6.tikz}
    \end{subfigure}
    ~
    \begin{subfigure}{0.4\linewidth}
	    \input{supp/mass_map_w10.tikz}
    \end{subfigure}
    \caption{Colormaps of step distance $S$ versus mass distribution for various frequencies -- Distal-to-center links' mass ratio $\frac{1}{2}(m_1+m_2)/m_0$ versus asymmetry ratio $m_1/m_2$}
    \label{fig:mass_map}
\end{figure*}

\begin{figure}
    \centering
    \begin{subfigure}{0.9\linewidth}
	    \input{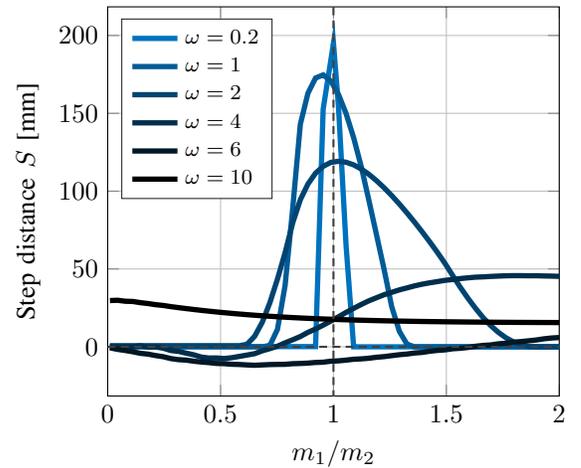}
    \end{subfigure}
    \caption{Step distance versus mass asymmetry $m_1/m_2$ for various frequencies $\omega$ and constant $\frac{1}{2}(m_1+m_2)/m_0=0.5$}
    \label{fig:S_gamar}
\end{figure}

In Fig.\,\ref{fig:mass_map} we present complete colormaps of the step distance $S$ the versus mass distribution in both parameters -- ratio of distal to center links' mass and asymmetry among the distal links. We also keep fixed amplitude $A=18^{\circ}$ and nominal angle $\varphi_0=110^{\circ}$, in order to investigate only the mass distribution. Fig.\,\ref{fig:S_gamar} depicts the effect of asymmetry in mass distribution among the distal links $m_1/m_2$, for various frequencies and constant ratio to the central link $\frac{1}{2}(m_1+m_2)/m_0=0.5$. Note that these curves are actually a section of the colormaps for selected frequencies.

\subsection*{Feedback control with hybrid inverse dynamics}
Assuming we can measure the whole state $\mathbf{q},\dot{\mathbf{q}}$, including contacts' velocities $\dot{x}_i$, which allows to deduce the contacts states, we propose the following control strategy: First, we now prescribe $\ddot{\mathbf{q}}_\text{c}(t,\mathbf{q},\dot{\mathbf{q}})$ as the controlled input (instead of prescribing $\mathbf{q}_\text{c}(t)$ and its derivatives as before). We calculate an ``ideal'' reference trajectory of the contacts velocities $\mathbf{v}_x^r(t)=[\dot{x}_1^r(t)\quad \dot{x}_2^r(t)]^\text{T}$ in which each contact slips for half of the cycle in the advancement direction only, and which corresponds to the reference joints trajectories $\mathbf{q}_\text{c}^r(t)$ in (\ref{eq:ref}). As long as the contacts states are stick-slip or slip-stick, the contacts velocities $\mathbf{v}_x^r(t)$ are maintained kinematically from the joint trajectories $\mathbf{q}_\text{c}^r(t)$. In these states we utilize PD reference tracking control law \cite{murray1994mathematical}. When slip-slip state is detected, it indicates undesired slippage. The control law is then \textit{switched} to PD reference tracking of $\mathbf{v}_r^x(t)$, translated with inverse dynamics \cite{murray1994mathematical} to the joints' inputs $\ddot{\mathbf{q}}_\text{c}(t)$ (detailed derivation follows). When trying to enforce $\mathbf{v}_r^x(t)$, this control law often results in limit configuration, causing ground-collision or tipping over. Hence another switch is introduced -- if either angle exceeds a predefined limit $\varphi_i\in [\varphi_{min},\varphi_{max}]\equiv \Phi$, the control law changes back to reference tracking of $\mathbf{q}_\text{c}(t)^r$. Overall the control law is
\begin{equation}\label{eq:control}
    \mathbf{\ddot{q}}_\text{c}=
    \left\{
    \begin{array}{lc}
    \multirow{2}{*}{$\mathbf{S}_1^{-1}\left[ \mathbf{\dot{v}}_x^r -\mathbf{K}_v (\mathbf{v}_x-\mathbf{v}_x^r)-\mathbf{S}_0 \right]$} & \text{if slip-slip}\\
    &  \&\ \varphi_i\in \Phi \\
    \mathbf{\ddot{q}}_\text{c}^r-\mathbf{K}_D (\mathbf{\dot{q}}_\text{c}-\mathbf{\dot{q}}_\text{c}^r)-\mathbf{K}_P (\mathbf{q}_\text{c}-\mathbf{q}_\text{c}^r) & \quad\text{otherwise}
    \end{array}
    \right.
\end{equation}
where $\mathbf{K}_v$, $\mathbf{K}_P$, $\mathbf{K}_D$ are $2\times 2$ positive-definite matrices of control gains $\mathbf{S}_0(\mathbf{q},\dot{\mathbf{q}})$, $\mathbf{S}_1(\mathbf{q})$ are inverse dynamics matrices derived next.

We consider a constrained state equation like (\ref{eq:full_b}), for slip-slip contacts state
\begin{multline}\label{eq:full_ss}
    \begin{bmatrix}\mathbf{M}_\text{b} & -\mathbf{E} & -\mathbf{W}_1\!^\text{T}\boldsymbol{\Gamma}_1 & -\mathbf{W}_2\!^\text{T}\boldsymbol{\Gamma}_2\\
    \{\mathbf{w}_1^y\}_\text{b} & 0\quad 0 & 0\quad 0 & 0\\
    \{\mathbf{w}_2^y\}_\text{b} & 0\quad 0 & 0\quad 0 & 0 \\
    \end{bmatrix}
    \begin{bmatrix}\ddot{\mathbf{q}}_\text{b} \\ \boldsymbol{\tau} \\ f_1^y \\ f_2^y\end{bmatrix}=
    \\
    =\begin{bmatrix}-\mathbf{M}_\text{c}\\
    -\{\mathbf{w}_1^y\}_\text{c}\\
    -\{\mathbf{w}_2^y\}_\text{c}\end{bmatrix}\ddot{\mathbf{q}}_\text{c}+
    \begin{bmatrix}-\mathbf{B}-\mathbf{G}\\
    -\dot{\mathbf{w}}_1^y\dot{\mathbf{q}}\\ -\dot{\mathbf{w}}_2^y\dot{\mathbf{q}}
    \end{bmatrix},
\end{multline}
which can be denoted as $\mathbf{A}(\mathbf{q})\mathbf{z}=\mathbf{b}(\mathbf{q},\dot{\mathbf{q}})+\mathbf{c}(\mathbf{q})\ddot{\mathbf{q}}_\text{c}$, where $\ddot{\mathbf{q}}_\text{b}=[\mathbf{I}_{3\times 3}\ 0_{3\times 5}]\mathbf{z}\equiv \mathbf{T}\,\mathbf{z}$. From (\ref{eq:full_ss}) the body coordinates' accelerations $\ddot{\mathbf{q}}_\text{b}$ can be found for prescribed $\ddot{\mathbf{q}}_\text{c}$ and known state as
\begin{equation}\label{eq:ddq_b}
    \ddot{\mathbf{q}}_\text{b}=\mathbf{T}\,\mathbf{A}^{-1}(\mathbf{q})\big[\mathbf{b}(\mathbf{q},\dot{\mathbf{q}})+\mathbf{c}(\mathbf{q})\ddot{\mathbf{q}}_\text{c}\big].
\end{equation}
Next, the contacts velocities are written in terms of the the Jacobian
\begin{equation}
    \mathbf{v}_x=\begin{bmatrix}\dot{x}_1\\ \dot{x}_2\end{bmatrix}=\begin{bmatrix}\mathbf{w}_1^x\\ \mathbf{w}_2^x\end{bmatrix}\dot{\mathbf{q}}\equiv \mathbf{W}^x\dot{\mathbf{q}}.
\end{equation}
This can be differentiated and decomposed into constrained- and body-coordinates as
\begin{equation}\label{eq:vx}
    \dot{\mathbf{v}}_x=\begin{bmatrix}\ddot{x}_1\\ \ddot{x}_2\end{bmatrix}=\dot{\mathbf{W}}^x\dot{\mathbf{q}}+\mathbf{W}^x\ddot{\mathbf{q}}=\dot{\mathbf{W}}^x\dot{\mathbf{q}}+\mathbf{W}^x_\text{b}\ddot{\mathbf{q}}_\text{b}+\mathbf{W}^x_\text{c}\ddot{\mathbf{q}}_\text{c}.
\end{equation}
Putting (\ref{eq:ddq_b}) into (\ref{eq:vx}) gives a relation between the joints' and the contacts' accelerations of the form $\dot{\mathbf{v}}_x~=~\mathbf{S}_1(\mathbf{q})\ddot{\mathbf{q}}_\text{c}~+~\mathbf{S}_0(\mathbf{q},\dot{\mathbf{q}})$, which defines the inverse dynamics matrices in the control law (\ref{eq:control}).

\subsubsection*{Proof of convergence}
The joints' reference tracking law can immediately be rearranged as $\ddot{\mathbf{e}}_\text{q}+\mathbf{K}_D\dot{\mathbf{e}}_\text{q}+\mathbf{K}_P\mathbf{e}_\text{q}=0$, where $\mathbf{e}_\text{q}(t)\equiv\mathbf{q}_\text{c}-\mathbf{q}_\text{c}^r$ is the joints' angles error, which we easily see to be converging (for positive-definite $\mathbf{K}_p$, $\mathbf{K}_D$). Putting the control law for contacts' velocities tracking (\ref{eq:control}) into (\ref{eq:vx}) with (\ref{eq:ddq_b}), also gives a converging velocities error $\mathbf{e}_v(t)\equiv \mathbf{v}_x-\mathbf{v}_x^r$ of the form $\dot{\mathbf{e}}_v+\mathbf{K}_v\mathbf{e}_v=0$. Of course this proof does not consider the the hybrid switching of the control law and the contact states, hence the convergence will only be local.

\begin{figure}
\centering
    \begin{subfigure}{0.9\linewidth}
	    \input{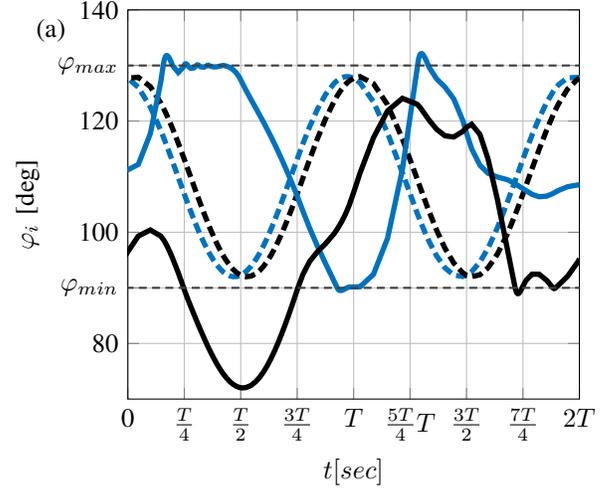}
    \end{subfigure}
    ~
    \begin{subfigure}{0.9\linewidth}
        \input{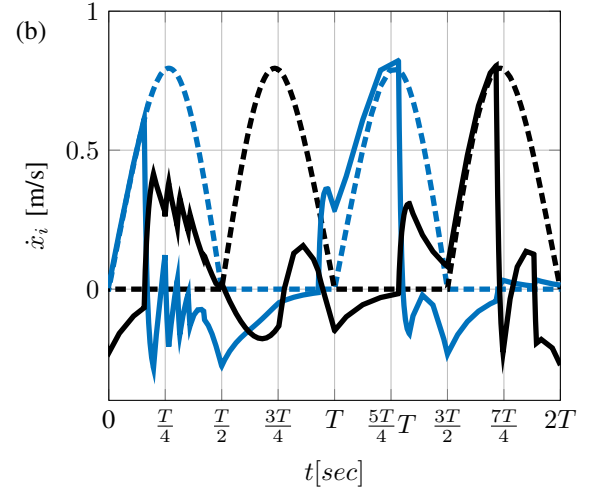}
	\end{subfigure}
\caption{Feedback control gait performance (solid curves) and reference trajectories (dashed curves) for $\omega=8$ [rad/s]. (a) Joint angles \textcolor{myblue}{$\varphi_1$ (blue)} and $\varphi_2$ (black). (b) Contacts' velocities \textcolor{myblue}{$\dot{x}_1$ (blue)} and $\dot{x}_2$ (black)}
\label{fig:control}
\end{figure}
~
\begin{figure}[H]
    \centering
    \begin{subfigure}{0.9\linewidth}
	    \input{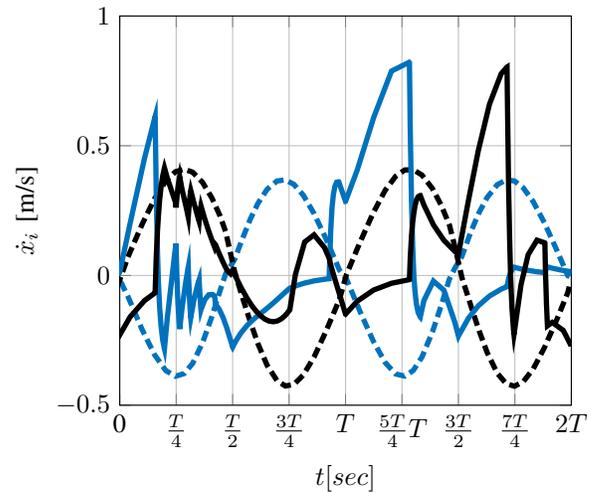}
    \end{subfigure}
    \caption{Contacts' velocities of feedback control gait (solid curves) and prescribed-angles gait (dashed curves) for $\omega=8$ [rad/s] -- \textcolor{myblue}{$\dot{x}_1$ (blue)} and $\dot{x}_2$ (black)}
    \label{fig:dx_control}
\end{figure}

An illustration of a gait with this control law, compared to the reference trajectories (in dashed curves) is given in Fig.\,\ref{fig:control} for control parameters $\mathbf{K}_v=10\mathbf{I}$, $\mathbf{K}_P=80\mathbf{I}$, $\mathbf{K}_D=100\mathbf{I}$, $\Phi=[90^{\circ},130^{\circ}]$, and reference joints trajectories (\ref{eq:ref}) with the parameters $\omega=8$~[rad/s], $A=18^{\circ}$, $\varphi_0=110^{\circ}$ and $\psi=20^{\circ}$. Interestingly, the solution converges in steady state to a two-step periodic solution, with mean distance $S=48.4$~[mm], compared to only 3.6 [mm] for the same joints trajectories without feedback. The difference is explained by comparing the the contacts' velocities with and without the feedback control in Fig.\,\ref{fig:dx_control}. Clearly, in this case the feedback control law significantly reduces the undesired opposing slippage, but such gaits were only found for a range of frequencies and careful tuneup of the control coefficients (as discussed in details in Section \ref{sec:control}, along with other limitations and requirements of this approach).

\end{document}